\documentclass[]{beingbeyond}
\usepackage{enumitem}
\usepackage[toc,page,header]{appendix}

\usepackage[utf8]{inputenc} % allow utf-8 input
\usepackage[T1]{fontenc}    % use 8-bit T1 fonts
\usepackage{hyperref}       % hyperlinks
\usepackage{url}            % simple URL typesetting
\usepackage{array}          % for advanced table column types
\usepackage{booktabs}       % professional-quality tables
\usepackage{amsfonts}       % blackboard math symbols
\usepackage{nicefrac}       % compact symbols for 1/2, etc.
\usepackage{microtype}      % microtypography
\usepackage{xcolor}         % colors
\usepackage{xspace}
\usepackage{bm}
\usepackage{bbm}
\usepackage{bbding}
\usepackage{tabularx}
\usepackage{textcomp}
\usepackage{amssymb}
\usepackage{enumitem}
\usepackage{amsmath}
\usepackage{mathtools}
\usepackage{multirow}
\usepackage{makecell}
\usepackage{color}
\usepackage{colortbl}
\usepackage{adjustbox}
\usepackage{caption}
\usepackage{graphicx}
\usepackage{wrapfig}
\usepackage{array}
\usepackage{multicol}
\usepackage{algorithm}
\usepackage{algorithmic}
\usepackage{diagbox}

\usepackage{mathptmx}  % 提供 Times 罗马体和配套数学字体
\DeclareFontFamily{T1}{georgiab}{}
\DeclareFontShape{T1}{georgiab}{m}{n}{<-> ptmr7t}{}
\DeclareFontShape{T1}{georgiab}{b}{n}{<-> ptmb7t}{}
\DeclareFontShape{T1}{georgiab}{m}{it}{<-> ptmri7t}{}
\DeclareFontShape{T1}{georgiab}{b}{it}{<-> ptmbi7t}{}
\definecolor{myyellow}{RGB}{255,192,0}
\definecolor{mygreen}{RGB}{107,170,64}
\definecolor{mywrite}{RGB}{255,227,132}

\definecolor{BlockC}{gray}{0.98}  
\definecolor{BlockA}{RGB}{191,211,230}
\definecolor{BlockB}{RGB}{199,233,192}

\title{Human-Centric Transferable Tactile Pre-Training \\ for Dexterous Robotic Manipulation}

\author{{\bfseries 
Chi Zhang$^{1,2,*}$ \quad 
Penglin Cai$^{1,2,*}$ \quad
Ziheng Xi$^{2,3}$ \quad
Haoqi Yuan$^{1,2}$ \\
Hao Luo$^{1,2}$ \quad
Wanpeng Zhang$^{1,2}$ \quad
Sipeng Zheng$^{2}$ \\
Chaoyi Xu$^{1,2}$ \quad
Zongqing Lu$^{1,2,\dagger}$
}}

\affiliation{{$^{1}$Peking University \quad $^{2}$BeingBeyond \quad $^{3}$Tsinghua University}}

\webpage{\url{https://beingbeyond.github.io/TTP/}}

\abstract{
    As an essential modality for dexterous and contact-rich tasks, tactile sensing provides precise force feedback that cannot be reliably inferred from vision. 
    However, limited by hardware and data collection systems, existing datasets with tactility remain small in scale and narrow in contact coverage. Meanwhile, Vision-Language-Action (VLA) models with tactile modality are constrained on dynamics-agnostic post-training, which limits the performance ceiling on downstream tasks.
    In this paper, we present H-Tac, a large-scale tactile-action dataset with 160-hour egocentric human videos containing more than 300 tasks and 135k episodes.
    Building upon this, we propose \textbf{T}ransferable \textbf{T}actile \textbf{P}re-Training (\textbf{TTP}), a system of tactile-based pre-training on human data for fine-grained robotic tasks.
    To bridge the gap between humans and robots, we use unified tactile and action spaces throughout the pre-training and post-training phases, preserving prior knowledge during human-to-robot transfer.
    By leveraging a tactile expert for future tactile prediction, our framework explicitly models the contact dynamics and precise physical interactions.
    Extensive experiments in simulation and on real robots demonstrate that our model achieves superior performance, exhibiting robust generalization and fine-grained manipulation capabilities.
    \textbf{TTP} paves the way for scalable tactile pre-training via human-to-robot transfer.
}

\checkdata[Date]{July 2, 2026}

\begin{document}

\maketitle
% 如果添加首图：需要将footnote移至\begin{document}上方，否则放在\maketitle下方
% ===== Footnote 补充信息 ====
\begingroup
\renewcommand\thefootnote{\fnsymbol{footnote}} % 使用自带符号编号: [1]* [2]† [3]‡
\setcounter{footnote}{0}
\footnotetext[1]{Equal contribution. Orders are decided by flipping a coin.}
\footnotetext[2]{Correspondence to Zongqing Lu $<$lu@beingbeyond.com$>$.}
\endgroup

\begin{figure}[htbp]
    \centering
    \includegraphics[width=1.0\textwidth]{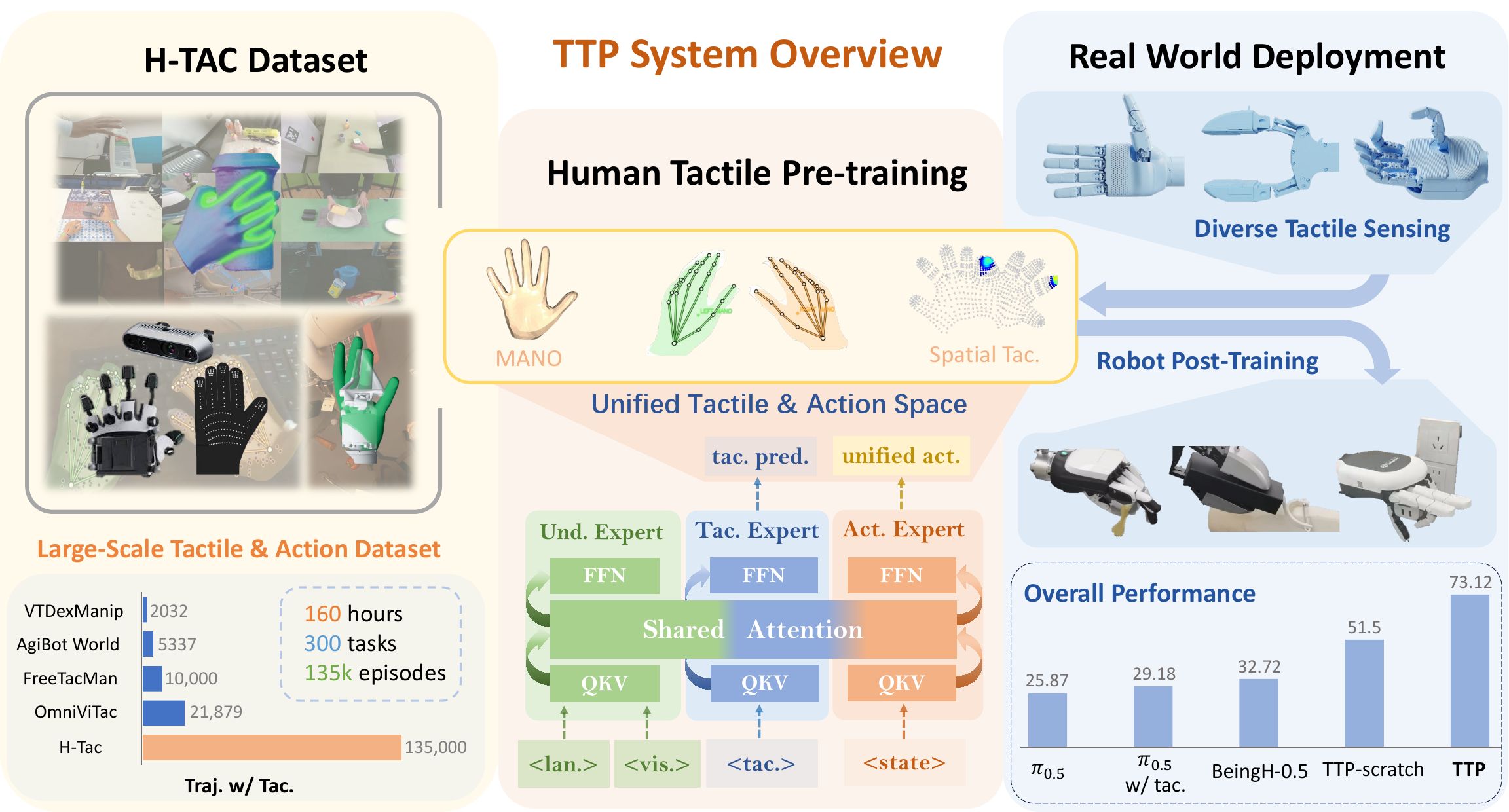}
    \caption{Overview of the Transferable Tactile Pre-Training \textbf{(TTP)} system.}
    \label{fig:teaser}
\end{figure}

\section{Introduction}

% tactile is import
% tactile data w/ teleoperation is of high cost
% thus using human data
% what architecture to leverge this human tactile data
% achieving human2robot transfer 
% dexterous -> fine-grained manipulation

While visual perception dominates in many robotic task, tactile sensing is essential for achieving complex, fine-grained, and dexterous manipulation,
especially when struggling with problems including occlusion and ambiguity in complex interactions. Tasks such as assembly, threading, or manipulating fragile items cannot be robustly executed without tactile sensing, indicating tactile sensing is a fundamental modality in advanced robotic systems. 
%Consequently, endowing embodied agents with tactile perception is critical for robust interaction with the physical world.

However, collecting tactile data on real robots remains difficult and expensive. Tactile sensors on different embodiments (especially dexterous hands) are non-unified in hardware integration, and teleoperating robots for contact-rich tasks are labor-intensive and hard to scale \cite{OSMO}. In contrast, acquiring human demonstration data is considerably easier and more scalable. This disparity has motivated a series of interest in learning from human demonstrations and human-to-robot skill transfer~\citep{luo2025beingh0, beingbeyond2026beingh05, zhang2025unitachand, yang2025egovla}. Yet, existing human demonstration datasets overwhelmingly focus on vision and action, largely overlooking the tactile modality. 
To fill in such gaps, collecting large-scale human-centric tactile-based dataset can be a possible solution.
% To fill in such gaps, we collect a large-scale, egocentric human demonstration dataset with synchronized tactile, visual, and action streams, laying the foundation for scalable human-centric tactile learning.

% Given such a tactile-rich human dataset, a pivotal question naturally arises: what model architecture can most effectively leverage this data? 
On the other hand, apart from tactile-rich human data, we also need proper architectures to learn skills and policies from these datasets.
Recent vision-language-action (VLA) models have exhibited impressive abilities in performing complex and long-horizon tasks, demonstrating strong capabilities of planning and semantically reasoning~\citep{kim2024openvla, black2024pi_0, intelligence2025pi05, beingbeyond2026beingh05}. In the meanwhile, learning from large-scale egocentric human videos has paving a promising path forward, as human data are relatively easy to collect and readily scalable~\citep{luo2025beingh0, beingbeyond2026beingh05, grauman2022ego4d, hoque2025egodex, yang2025egovla}. However, standard VLA models, which lack tactile sensing, suffer from a critical limitation: the learned policies remain predominantly driven by visual cues and systematically ignore tactile information, leading to degraded performance in contact-intensive scenarios. % Without tactile grounding, such models struggle to adapt to contact-rich scenarios with subtle physical interactions, especially with occlusion and ambiguity.
This leads to a natural yet challenging question: \emph{can we unify large-scale tactile pre-training within the VLA paradigm and form a transferrable tactile-based pre-training for human-to-robot skill transfer?}

To address this, we propose \textbf{T}ransferable \textbf{T}actile \textbf{P}re-Training (\textbf{TTP}), a system of human-centric tactile pre-training for transferable robot skill learning. Our system includes H-Tac, a human-centric tactile dataset (Section~\ref{sec:dataset}), tactile-based pre-training (Section~\ref{sec:pretrain}), as well as post-training for downstream tasks (Section~\ref{sec:experiments}).
We propose to pre-train a VLA model on large-scale egocentric human videos enriched with tactile and action data. This pre-training phase endows the model with rich domain-relevant priors and enables it to leverage the inherent alignment capabilities of VLA architectures to implicitly associate tactile signals with vision and language. The model is subsequently post-trained on downstream robotic tactile tasks while maintaining strict consistency with the pre-training setup, avoiding the pre-train/post-train distribution mismatch in previous literature~\citep{zhang2026vtla, li2026atvla, cheng2025omnivtla, bi2026vlatouch}. To preserve consistency between human pre-training and robot post-training, our model adopts a unified action space and a unified tactile space that standardizes heterogeneous tactile representations across embodiments. Additionally, we propose to build the model as a dual-expert system, in which an action expert generating future action chunks and a tactile expert predicting future tactile signals, which effectively models the tactile dynamics of the environment. These objectives encourage the model to balance semantic reasoning and physical interaction, bridging the long-standing gap between high-level task understanding and low-level fine-grained contact control.

Our contributions are threefold:
\begin{itemize}
    \item We collect and open-source a large-scale egocentric human demonstration dataset with dense tactile and action annotations and verify the benefit of such human-centric tactile data, which provides a critical resource for researches on tactile-conditioned pre-training. % 并且验证了这些数据集能够对human-to-robot有用
    % system: data collection
    \item We are the first to leverage tactile pre-training with human-centric demonstrations, enabling a VLA model to acquire tactile-grounded priors at scale before robot-specific post-training.
    \item After post-training on a suite of real-robot tactile tasks, our model achieves superior performance with precise and fine-grained manipulation, exhibiting strong cross-embodiment capabilities and demonstrating the effectiveness of human-to-robot policy transfer.
\end{itemize}

% contributions:
% 1. open-source tactile dataset
% 2. the first to introduce tactile pre-training with human-centric demonstrations
% 3. after post-training on tactile tasks, have a good performance 

\section{Related Work}

% Tactile-based manipulation (IL, RL, methods, datasets, benchmarks, etc.)

% Human2robot

% fusing tactile into VLA

\textbf{Tactile-based manipulation.} 
To achieve dexterous and fine-grained manipulation, there has been many works on tactile-relevant tasks, datasets, benchmarks, and policies in the previous literature.
Tactile-based datasets including OpenTouch~\citep{song2025opentouch}, EgoPressure~\citep{zhao2025egopressure}, VTDexManip~\citep{liu2025vtdexmanip} and OmniViTac~\citep{zheng2026omnivta} provide large-scale tactile-action aligned data for contact-rich tasks.
As a tactile benchmark, ROTO~\citep{miller2026enhancing} is proposed to encourage embodied agents to incorporate tactile sensing to overcome sensory deficits and reliance on idealised state information.
As for tactile-based policies, RDP~\citep{xue2025reactive} is proposed as a slow-fast system, with the fast policy takes tactile sensing as input for reactive responses. Other methods based on reinforcement learning (RL)~\citep{hu2025tactile} leverages binary contact information in decoupled robot-object motion.
In this paper, we tackle the problems in tactile-relevant manipulation by collecting human-centric transferable data and tactile-based pre-training, which demonstrates robust capabilities of fine-grained manipulation.

\textbf{Human to robot skill transfer.}
Due to the high cost of collecting robot data, many previous works opt for learning from human-centric skills or human demonstrations, transferring such learned prior knowledge in the human space into robot space.
Some works focus on learning robot arms and/or dexterous hands manipulation from demonstration teaching, teleoperation, or human videos~\citep{chi2024universal, bharadhwaj2025gen2act, xie2026human2robot, kim2025uniskill, zhou2026traj2action, heppert2026scaling}, while pre-training on large-scale egocentric human demonstration videos has become widely used~\citep{kareer2025egomimic, zheng2026egoscale, zhang2026unidex, luo2025beingh0, beingbeyond2026beingh05}.
In the field of tactile-based fine-grained manipulation, TactAlign~\citep{wi2026tactalign} and UniTacHand~\citep{zhang2025unitachand} pave the path towards a universal tactile representation aligning human hands and dexterous robotic hands, and UniTacHand demonstrates zero-shot human-to-robot transfer with only a small amount of paired data. 
In our work, we achieve such a transfer by human-tactile pre-training on a vision-language-action (VLA) model. After few-shot post-training on downstream tasks, our model demonstrates excellent performance with fine-grained manipulation and robust control.

\textbf{Fusing tactile modality into vision-language-action (VLA) models.}
Some literature explores to fuse tactile modality into VLA models, forming vision-tactile-language-action (VTLA) models.
Following TLA~\citep{hao2026tla}, VTLA~\citep{zhang2026vtla} is directly trained on simulation-collected data with both vision and visuo-tactile modalities on the basis of a vision-language model (VLM). Tactile-VLA~\citep{huang2025tactile} introduces tactile-aware instruction following, leveraging the language common senses acquired by the pre-training phase of VLA. 
% Some other works~\citep{zhang2026craft, li2026favla, li2026atvla} focus on the problem of force-vision mismatch in the semantic or time domain during post-training. CRAFT~\citep{zhang2026craft} uses force-aware curriculum fine-tuning to tackle the problem of over-relying on vision and ignoring force information, adapting VLA models to contact-rich manipulation. FAVLA~\citep{li2026favla} disentangles low-frequency semantic planning with high-frequency force-relevant control, forming a fast-slow system. Similarly, AT-VLA~\citep{li2026atvla} deploys a dual system with tactile injecting and reaction in the inference stream.
Some other works~\citep{zhang2026craft, li2026favla, li2026atvla} focus on the problem of tactile-vision mismatch in the semantic or time domain during post-training, adopting curriculum tactile fine-tuning or disentangling low-frequency semantic planning with high-frequency tactile-relevant control.
% Instead of directly injecting tactile tokens in VLA models with all the burden left on the post-training phase, some other works tend to use an explicit learning process to align tactility and other modalities. OmniVTLA~\citep{cheng2025omnivtla} provides a semantically-aligned tactile encoder by pre-training on manipulation data of various categories of objects, which serves as an excellent initialization for the VLA. From another perspective, VLA-Touch~\citep{bi2026vlatouch} chooses to freeze the VLA model, and use a pre-trained tactile-language model to provide semantic tactile feedback for high-level task planning, as well as another diffusion-based tactile-conditioned controller to refine actions generated by the VLA. HapticVLA~\citep{gubernatorov2026hapticvla} opts for a teacher-student distillation paradigm, in which the tactile-relevant reward-weighted expert is distilled to a tactile-agnostic conventional VLA.
Instead of directly injecting tactile tokens in VLA models with all the burden left on the post-training phase, some other works~\citep{cheng2025omnivtla, bi2026vlatouch, gubernatorov2026hapticvla, zhao2026fdvla} tend to use an explicit learning process to align tactility and other modalities before plugging the relevant modules into policies.
In our work, we lead in tactile-included pre-training, making the implicit tactile fusing process much more efficient, adaptive and generalizable with pre-trained prior knowledge.

\section{Tactile-Based Dataset Collection for Pre-Training}
\label{sec:dataset}

\begin{figure}[t]
    \centering
    \includegraphics[width=0.98\textwidth]{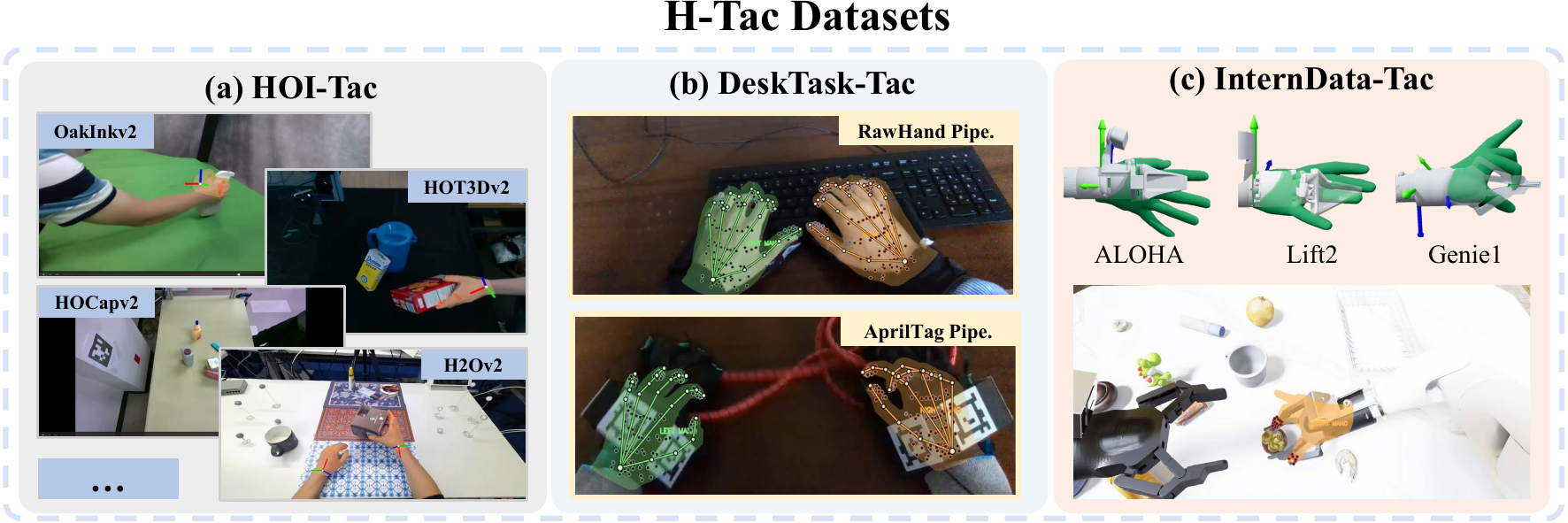}
    \caption{Our H-Tac datasets, composed of (a) HOI-Tac, (b) DeskTask-Tac, and (c) InternData-Tac. In total, H-Tac contains 160-hour vison-tactile-action data, including 300+ tasks and 135k+ episodes.}
    \label{fig:datasets}
\end{figure}

In this section, we introduce our tactile-based dataset for pre-training, overviewed as Figure~\ref{fig:datasets}.

\subsection{HOI-Tac Dataset}
We use multiple public datasets spanning hand-object (ARCTIC~\cite{fan2023arctic}, DexYCB~\cite{chao2021dexycb}, H2O~\cite{kwon2021h2o}, H2O3D~\cite{hampali2022keypoint}, HO3D v2/v3~\cite{hampali2020honnotate}, HOCap~\cite{wang2025hocap}, HOI4D~\cite{liu2022hoi4d}, HOT3D~\citep{banerjee2025hot3d}, InterHand2.6M~\citep{moon2020interhand2}, OakInk-v1~\cite{yang2022oakink}, OakInk-v2~\cite{zhan2024oakink2}), hand-face (Decaf~\cite{brahmbhatt2024decaf}), and hand-scene (PROX~\cite{hassan2019resolving}, RICH~\cite{huang2022rich}) interactions.
For each frame, we generate per-vertex binary contact labels on the 778-vertex MANO hand mesh by thresholding the distance between the hand surface and object meshes.
These per-vertex contact signals are then projected onto the 351-taxel UniTacHand UV space~\cite{zhang2025unitachand} to serve as tactile supervision, forming our HOI-Tac dataset.
In total, the composite contains approximately 11.5M frames ($\sim$106 hours) across 124.8K sequences, encompassing egocentric videos with single-hand and bimanual grasps, static and dynamic object interactions, and diverse environments from tabletop to whole-body scenes.

\subsection{DeskTask-Tac Dataset}

\begin{figure}[htbp]
    \centering
    \includegraphics[width=0.8\textwidth]{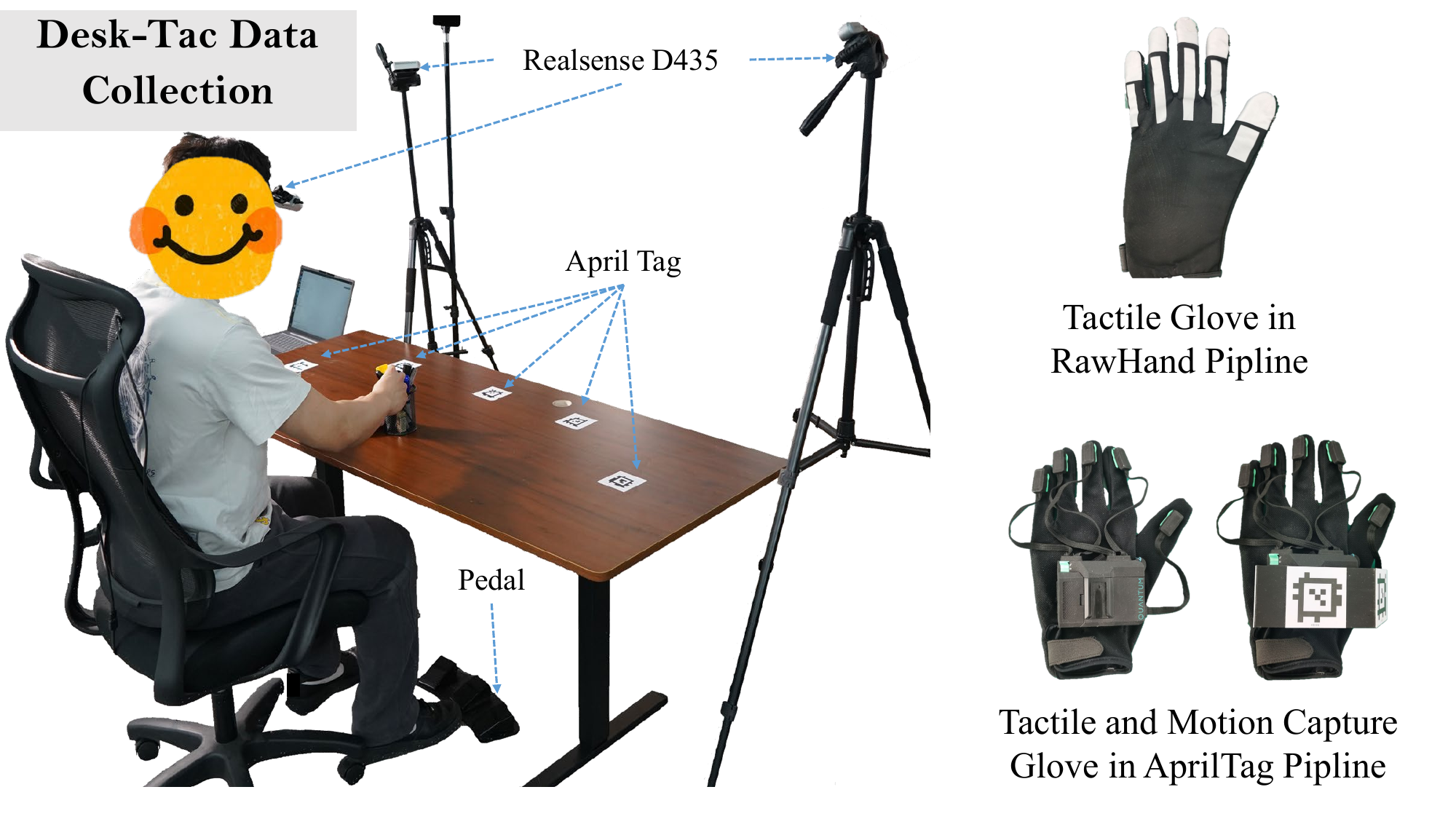}
    \caption{Data collection system of our DeskTask-Tac dataset.}
    \label{fig:data-collection-system}
\end{figure}

We design a data collection system as in Figure~\ref{fig:data-collection-system} to collect our DeskTask-Tac dataset, which is for bimanual manipulation data in real-world desktop scenarios. Three RealSense cameras, including two external-view devices and an egocentric one, record the videos in different views. Each episode uses the first-person video timeline as the main axis, organizing hand geometry, tactile data, task labels, and action targets. The system supports two types of upstream hand reconstruction pipelines:

\textbf{RawHand Pipeline}: We use image-based hand keypoint detection and multi-view triangulation to recover the 3D state of the hands, reducing reliance on external reference hardware on hands. In this pipeline, we only need a tactile glove to record the tactility on the human hands.

\textbf{AprilTag Pipeline}: We use scene AprilTags to recover bimanual poses combining hand reference structures in MANO \cite{romero2022mano} and tactile information. In this case, we need both a tactile and a motion capture (MoCap) glove to record the hand motion and tactility simultaneously, with additional april tags for wrist locating.

%The raw session initially saves multiple RealSense video streams, timestamps, binary tactile streams, task metadata, and raw hand states. 
In the post-processing phase, we use the first-person camera timeline as a reference to perform temporal sampling or interpolation on multi-view observations, tactile readings, and hand states.
The processed Spatial Tactile representation combines tactile signals, MANO parameters~\citep{romero2022mano}, and the wrist pose to estimate the 3D coordinates for each tactile unit. Given the hand pose $\theta_h \in \mathbb{R}^{45}$, shape $\beta_h \in \mathbb{R}^{10}$, and the wrist pose in the camera frame $(R_h,t_h)$, the MANO model yields 778 mesh vertices and 21 joints. The tactile sensors are mapped to an 891-dimensional UV vertex, then converted to 351-dimensions via validation mask.
In total, the DeskTask-Tac dataset contains 37.2 hours of 30 Hz data (947 episodes, $\sim$4M frames).
%Finally, a rigid transformation is applied to transform them into the first-person camera coordinate system:
%\begin{equation}
%  \mathbf{p}^{c}_{h,i} = R_h \left(\sum_{j=1}^{891} S_{ij}\,\mathbf{u}_{h,j}(\theta_h,\beta_h)\right) + t_h,
%  \quad i=1,\ldots,137 .
%\end{equation}

\subsection{Tactile-Augmented InternData-Tac Dataset}

We augment the InternDataEngine \cite{tian2025interndata} pipeline with a lightweight tactile recorder that saves contact forces and patch details (position, normal, distance, force) as an episodic sidecar, strictly preserving the original visuo-linguo-motor streams. To enable cross-embodiment supervision independent of specific robot grippers, we project the collected contact patches onto a shared MANO hand surface. 
%Given the unified end-effector (EEF) to world transform $T^{W}_{U}(t)$ and a calibrated MANO root to EEF transform $T^{U}_{M}$, the MANO-to-world alignment is:
%$$T^{W}_{M}(t) = T^{W}_{U}(t) T^{U}_{M}$$
Active contact patches are transformed into the MANO frame, dispersed to adjacent vertices via a Gaussian kernel, and converted to local vertex pressure to yield a compact 351-D tactile vector. Additionally, near-surface zero-force patches are converted into a distance-decayed pseudo-contact signal, providing geometric proximity supervision prior to physical impulses.
%The released dataset adopts a LeRobot-style layout, comprising first-person video, language instructions, EEF poses, MANO parameters, and normalized ($[0,1]$) tactile fields (\texttt{tactile.\{left,right\}\_mano\_tactile}). 
Inactive arm fields are ignored to prevent confusion with valid zero-contact states. In total, the dataset encompasses 17.8 hours of 30 Hz data (9,563 episodes, $\sim$1.9M frames) across three diverse robot configurations: Genie1, Lift2, and Split ALOHA.

\subsection{Dataset Statistics}

\begin{figure}[htbp]
    \centering
    \subfloat[The mean tactile values visualized on MANO surfaces (left and right hands).]{
        \includegraphics[width=0.47\textwidth]{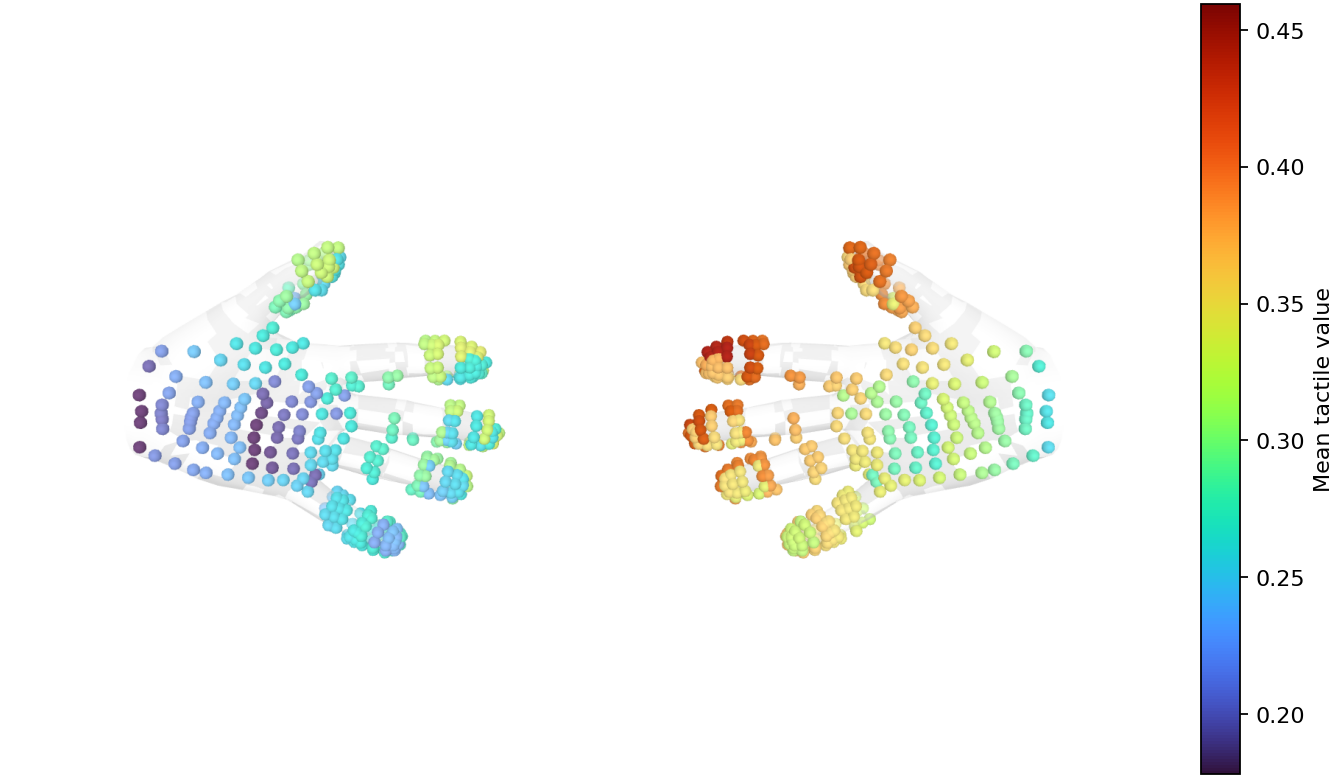}
    }
    \hfill
    \subfloat[Statistics on task language instruction prefixes (sorted by count).]{
        \includegraphics[width=0.47\textwidth]{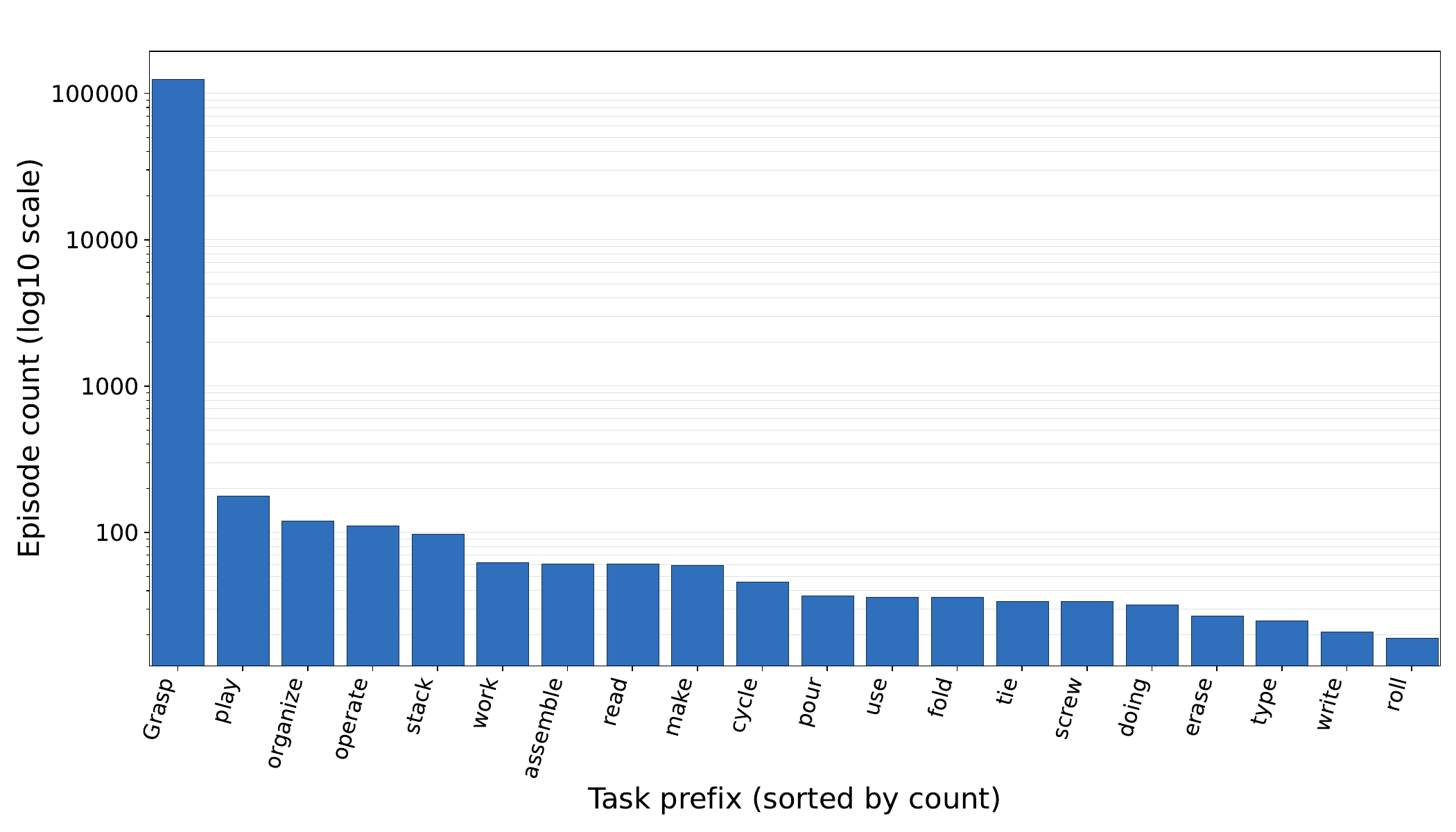}
    }
    \caption{Statistics on our pre-training datasets.}
    \label{fig:dataset-statistics}
\end{figure}

We provide some statistics over our pre-training datasets, as shown in Figure~\ref{fig:dataset-statistics}. Specifically, we provide:
\begin{enumerate}[label={(\alph*)}]
    \item The mean tactile values visualized on MANO surfaces (left and right hands). 
    \item Statistics on task language instruction prefixes (sorted by count).
\end{enumerate}

Figure~\ref{fig:dataset-statistics} (a) shows that right hands have tactile sensings with higher magnitudes overall, and tactile readings on fingertips are much more significant than those on palms, indicating that hand-object contacts on fingertips appear more frequently than on palms in these datasets.

From Figure~\ref{fig:dataset-statistics} (b), we can see that most language instructions begin with verbs, and the word ``grasp'' occupies an absolute dominant position compared with other verbs (e.g., stack, pour, erase).

% \begin{figure}
%     \centering
%     \begin{minipage}{0.45\textwidth}
%         \includegraphics[width=\textwidth]{images/tactile-mean.png}
%         \captionof{subfigure}{...}
%     \end{minipage}
%     \hfill
%     \begin{minipage}{0.45\textwidth}
%         \includegraphics[width=\textwidth]{images/task_prefix_counts.pdf}
%         \captionof{subfigure}{...}
%     \end{minipage}
%     \caption{Caption}
% \end{figure}

\section{Tactile-Based Pre-Training}
\label{sec:pretrain}

Our model is built on top of BeingH-0.5~\citep{beingbeyond2026beingh05}, a state-of-the-art foundation VLA model with unified cross-embodiment control capabilities.
BeingH-0.5 contains a multimodal understanding expert initialized from InternVL-3.5~\citep{wang2025internvl3.5} and an action generation expert for robot control.
We extend the VLM part to tactile modality, and introduce a novel tactile prediction expert, enabling the model to jointly predict future actions and future tactile information .

\subsection{Problem Formulation}

We consider table-top fine-grained manipulation tasks with multi-modal observations.
At physical timestep $t$, the observation consists of a language instruction $l$, one or multiple RGB images
$\mathbf{I}_t=\{\mathbf{I}_t^{(v)}\}_{v=1}^{V}$ with
$\mathbf{I}_t^{(v)}\in\mathbb{R}^{H_I\times W_I\times 3}$ from different views $v$, proprioceptive state
$s_t\in\mathbb{R}^{D_{\mathrm{act}}}$, and tactile readings
$o_t\in\mathbb{R}^{D_{\mathrm{tac}}}$.
Here $H_I$ and $W_I$ denote image height and width, $D_{\mathrm{act}}$ denotes the state and action space dimension, and $D_{\mathrm{tac}}$ denotes the tactile space dimension.
We use $K$ to denote the prediction horizon, i.e., the action chunk length.

To preserve recent contact information, the policy conditions on a strided tactile history:
\begin{equation}
    \mathcal{O}^{\mathrm{hist}}_t
    =
    \left[
    o_{t-d(L-1)}, \ldots, o_{t-d}, o_t
    \right]
    \in \mathbb{R}^{L\times D_{\mathrm{tac}}},
\end{equation}
where $L$ is the history length and $d$ is the temporal stride, balancing between reducing the computational burden and preserving dominant tactile information among frames.
The model predicts both a future action chunk and future tactile readings:
\begin{equation}
    A_t =
    [a_t,\ldots,a_{t+K-1}]
    \in \mathbb{R}^{K\times D_{\mathrm{act}}},
    \qquad
    O_t^{+} =
    [o_t,\ldots,o_{t+K-1}]
    \in \mathbb{R}^{K\times D_{\mathrm{tac}}}.
\end{equation}
Thus, the policy is formulated as
\begin{equation}
\label{eq:formulation}
    \left(
    \hat{A}_t,\hat{O}_t^{+}
    \right)
    \sim
    \pi_\theta
    \left(
    A_t,O_t^{+}
    \mid
    l,\mathbf{I}_t,s_t,\mathcal{O}^{\mathrm{hist}}_t
    \right).
\end{equation}

\begin{figure}[t]
    \centering
    \includegraphics[width=1.0\textwidth]{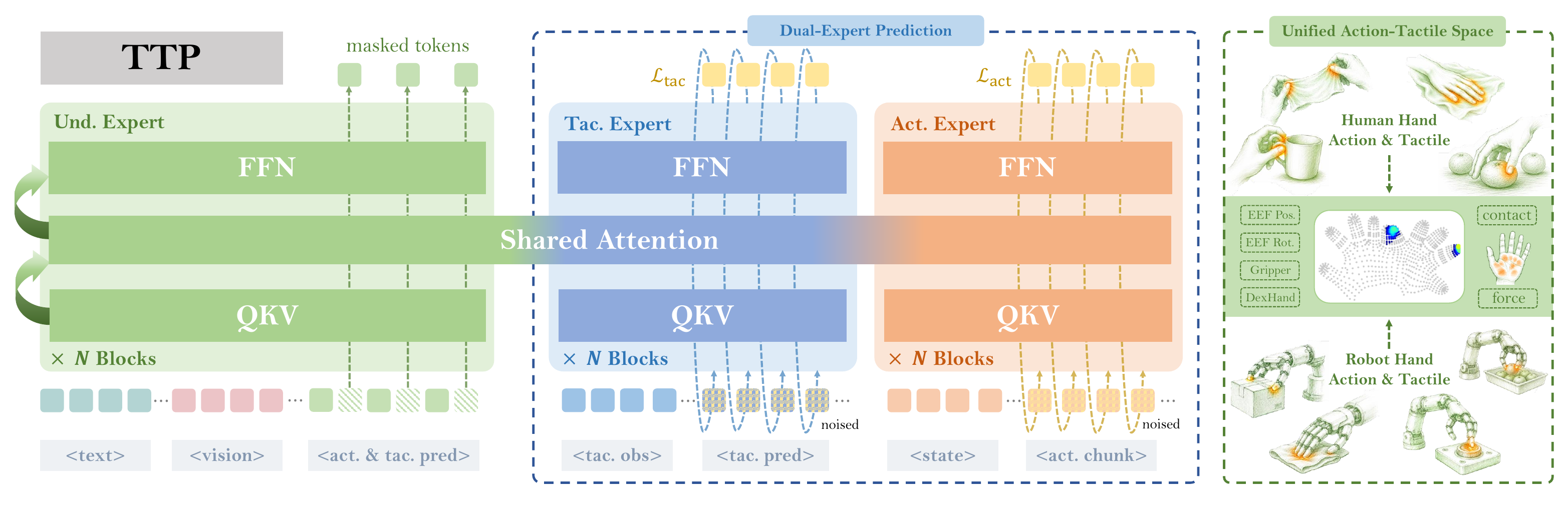}
    \caption{Training architecture of TTP. Our model includes an understanding expert for visual and text interpretation, an action expert, and a tactile expert. We use a unified action and tactile space to preserve pre-traing period knowledge.}
    \label{fig:architecture}
\end{figure}

\subsection{Unified Action and Tactile Space}

We aim to keep the consistency between human-demonstrations pre-training and real-robot post-training, with various embodiments including human hands, dexterous hands with piezoresistive tactile sensings, and even grippers with visuo-tactile sensings. Such a unification requires a universal action and tactile space to represent different actions and tactile readings on different embodiments while preserving morphological structures and meanings. 

Following BeingH-0.5~\citep{beingbeyond2026beingh05}, our unified action space contains $D_{\mathrm{act}}=200$ dimensions, including end effector pose (location and axis-angle rotation), dexterous hand actions, human MANO values (beta, translocations, and theta)~\citep{romero2022mano}, etc.

The 200-dimensional space is semantically organized into slots as in Table~\ref{tab:unified-action-space}, with each slot has a specific purpose.

\begin{table}[t]
\centering
\caption{Semantic organization of the 200-dimensional action space.}
\label{tab:unified-action-space}
\resizebox{0.9\textwidth}{!}{
\begin{tabular}{cccc}
\toprule
\textbf{Slot Indices} & \textbf{Semantic Name} & \textbf{Dimensions} & \textbf{Description} \\
\midrule
\rowcolor{BlockA!20} \multicolumn{4}{l}{\textbf{Right Arm End-Effector (Dims 0-8)}} \\
% \midrule
0-2 & eef\_position & 3 & Right arm end-effector position (x, y, z) \\
3-5 & eef\_rotation & 3 & Right arm end-effector rotation (axis-angle) \\
6-8 & Reserved & 3 & Reserved for future use \\
\midrule
\rowcolor{BlockB!20} \multicolumn{4}{l}{\textbf{Left Arm End-Effector (Dims 9-17)}} \\
% \midrule
9-11 & left\_eef\_position & 3 & Left arm end-effector position (x, y, z) \\
12-14 & left\_eef\_rotation & 3 & Left arm end-effector rotation (axis-angle) \\
15-17 & Reserved & 3 & Reserved for future use \\
\midrule
\rowcolor{BlockA!20} \multicolumn{4}{l}{\textbf{Grippers (Dims 18-19)}} \\
% \midrule
18 & gripper\_position & 1 & Right gripper open/close (0=closed, 1=open) \\
19 & left\_gripper\_position & 1 & Left gripper open/close \\
\midrule
\rowcolor{BlockB!20} \multicolumn{4}{l}{\textbf{Dexterous Hands (Dims 20-43)}} \\
% \midrule
20-25 & dexhand\_position & 6 & Right dexterous hand joints \\
26-31 & Reserved & 6 & Right hand extension \\
32-37 & left\_dexhand\_position & 6 & Left dexterous hand joints \\
38-43 & Reserved & 6 & Left hand extension \\
\midrule
\rowcolor{BlockA!20} \multicolumn{4}{l}{\textbf{Legacy/Special Slots (Dims 44-49)}} \\
% \midrule
44-45 & libero\_gripper\_position & 2 & LIBERO-specific gripper state \\
46-49 & Reserved & 4 & Reserved for future use \\
\midrule
\rowcolor{BlockB!20} \multicolumn{4}{l}{\textbf{Arm Joints (Dims 50-69)}} \\
% \midrule
50-56 & arm\_joint\_position & 7 & Right arm joint positions (7-DoF) \\
57-63 & left\_arm\_joint\_position & 7 & Left arm joint positions (7-DoF) \\
64-65 & head\_position & 2 & Head pan/tilt joints \\
66-68 & waist\_position & 3 & Waist/torso joints \\
69 & Reserved & 1 & Reserved for future use \\
\midrule
\rowcolor{BlockA!20} \multicolumn{4}{l}{\textbf{Mobile Base (Dims 70-75)}} \\
% \midrule
70-72 & base\_position & 3 & Mobile base position (x, y, z) \\
73 & base\_motion & 1 & Base motion command \\
74 & control\_mode & 1 & Control mode flag \\
75 & Reserved & 1 & Reserved for future use \\
\midrule
\rowcolor{BlockB!20} \multicolumn{4}{l}{\textbf{Reserved (Dims 76-89)}} \\
% \midrule
76-89 & Reserved & 14 &  Reserved for future embodiments and extensions \\
\midrule
\rowcolor{BlockA!20} \multicolumn{4}{l}{\textbf{Human Hands (Dims 90-199)}} \\
% \midrule
90-99  & right\_beta  & 10 & Right Hand Shape (only for state, MANO parameter $\beta$)  \\
100-109 & left\_beta   & 10 & Left  Hand Shape (only for state, MANO parameter $\beta$)  \\
110-154 & right\_theta & 45 & Right Hand Articulation (axis-angle, MANO parameter $\theta$) \\
155-199 & left\_theta  & 45 & Left  Hand Articulation (axis-angle, MANO parameter $\theta$) \\
\bottomrule
\end{tabular}
}
\end{table}

% Most of our training datasets (all of the pre-training human demonstrations as well as most of the real-robot experimental datasets) fall in the field of piezoresistive tactilities on (human and/or robot) hands, and 
We use UniTacHand~\citep{zhang2025unitachand} as the unified space of tactile representation to yield a maximum preservation of morphological consistency. Following UniTacHand, we preserve $D_{\mathrm{tac}}=351$ taxels for each hand, which are distributed on the surface of the MANO hand model. We project piezoresistive tactilities from different embodiments onto these taxels, which preserves the consistency between pre-training (human hands) and post-training (robot hands). Similar to the unified action space in BeingH-0.5, such a unified tactile space enables our model with capabilities of cross-embodiment tactile prediction, exhibiting strong performances in various arm-hand combinations.

\subsection{Large Scale Tactile Pre-Training}

We represent supervision as a unified multimodal sequence and train the model in a VQA-style query-answer format
$[\mathcal{S}_Q;\mathcal{S}_A]$.
The query $\mathcal{S}_Q$ contains image tokens, language tokens, proprioceptive state tokens, and tactile observation tokens, while the answer $\mathcal{S}_A$ contains action tokens and tactile prediction tokens.

Let $H_t$ denote the observation-conditioned token-level context at physical timestep $t$.
During flow matching, after inserting the noisy action or tactile trajectory at flow timestep $\tau$, the understanding expert produces a hidden context denoted by
$H_{t,\tau}$.
Here, $H_t$ is the observation context, while $H_{t,\tau}$ is the flow-time-dependent context used for velocity prediction.

For each modality $m\in\{\mathrm{act},\mathrm{tac}\}$, define the clean target as
$x_1^{\mathrm{act}} = A_t$, and $x_1^{\mathrm{tac}} = O_t^{+}$,
we sample $x_0^m\sim\mathcal{N}(\mathbf{0},\mathbf{I})$ and a flow timestep $\tau^m\in[0,1]$, then construct
\begin{equation}
    x_{\tau^m}^{m}
    =
    (1-\tau^m)x_0^m + \tau^m x_1^m.
\end{equation}
The target velocity is
\begin{equation}
    u^m = x_1^m - x_0^m.
\end{equation}
The action expert and tactile expert predict
\begin{equation}
    \hat{u}_\theta^m
    =
    v_\theta^m
    \left(
    x_{\tau^m}^{m}, \tau^m, H_{t,\tau^m}
    \right),
    \qquad
    m\in\{\mathrm{act},\mathrm{tac}\}.
\end{equation}

Therefore, the flow matching loss is
\begin{equation}
\label{eq:flow-matching-loss}
    \mathcal{L}_{m}
    =
    \mathbb{E}_{x_0^m,\tau^m}
    \left[
    \left\|
    \left(
    v_\theta^m(x_{\tau^m}^{m},\tau^m,H_{t,\tau^m})
    -
    (x_1^m-x_0^m)
    \right)
    \right\|_2^2
    \right],
    \qquad
    m\in\{\mathrm{act},\mathrm{tac}\},
\end{equation}
after masking padded or unavailable dimensions.
The total objective is
\begin{equation}
\label{eq:total_loss}
    \mathcal{L}
    =
    \lambda_{\mathrm{act}}\mathcal{L}_{\mathrm{act}}
    +
    \lambda_{\mathrm{tac}}\mathcal{L}_{\mathrm{tac}},
\end{equation}
where $\lambda_{\mathrm{act}}$ and $\lambda_{\mathrm{tac}}$ are the weights of each loss term.

\subsection{Tactile-Action Manifold-Preserving Gating}

\begin{figure}[t]
    \centering
    \includegraphics[width=0.68\textwidth]{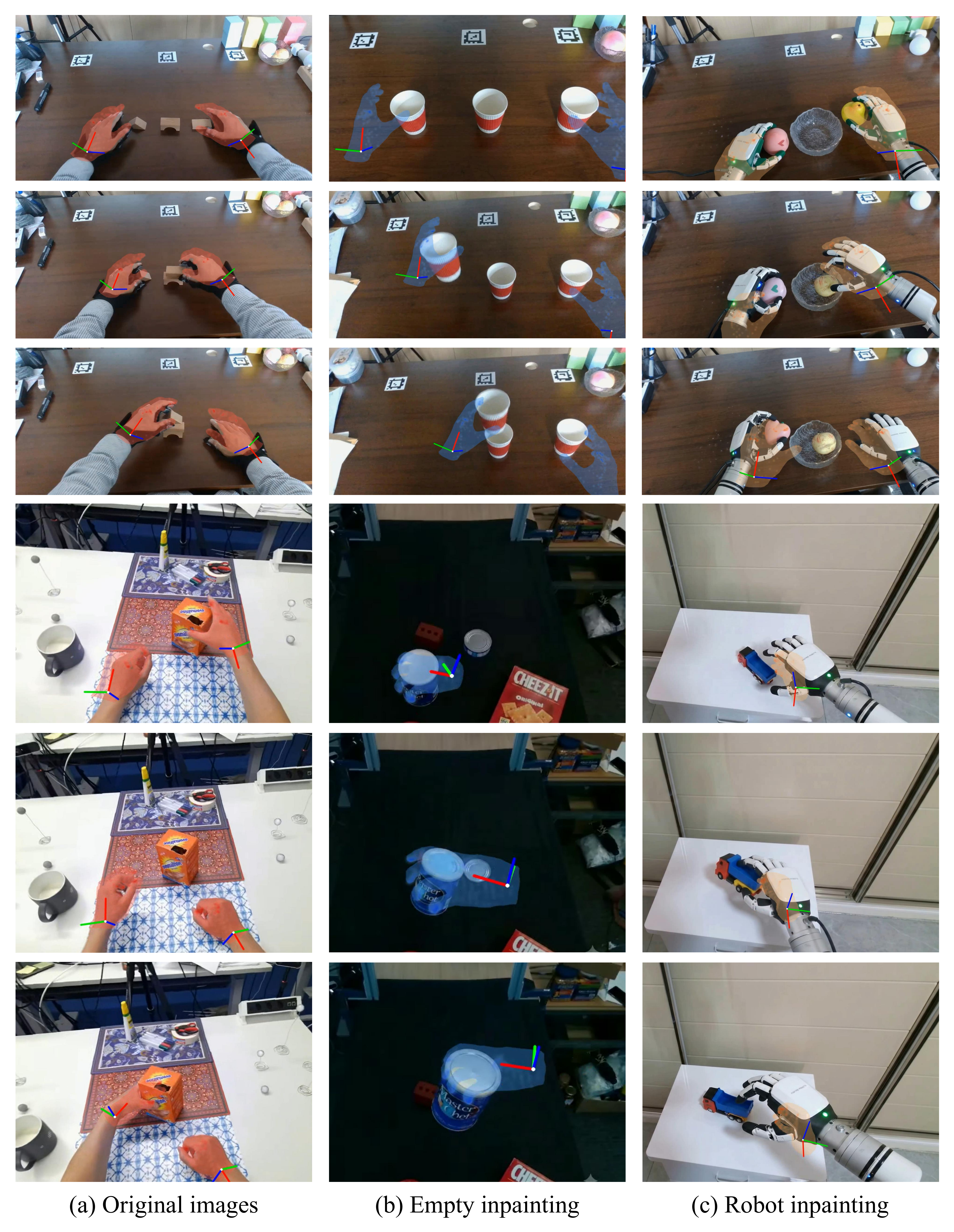}
    \caption{Visualization showcase. After tactile-based pre-training, our TTP model can generate hand motion and tactile predictions well, and can generalize to OOD inpainted scenes.}
    \label{fig:visualization}
\end{figure}

Tactile-Action Manifold-Preserving Gating (MPG) operates on the same flow-time-dependent context $H_{t,\tau}$ used by the action and tactile experts.
Specifically, $H_{t,\tau}$ denotes the suffix token features projected to the VLM hidden space, including proprioceptive state, tactile observation, noisy action, and noisy tactile prediction token features at physical timestep $t$ and flow timestep $\tau$.
MPG enhances this context before velocity decoding by $\tilde{H}_{t,\tau} = \mathrm{MPG}(H_{t,\tau})$, with the specific formula in Equation~\ref{eq:mpg}.

The velocity field is then evaluated as
\begin{equation}
    \hat{u}_\theta^m
    =
    v_\theta^m
    \left(
    x_\tau^m,\tau,\tilde{H}_{t,\tau}^{m}
    \right).
\end{equation}

Since $H_{t,\tau}$ contains task-specific and behavior-specific semantics, small distribution shifts in the conditioning context may lead to unstable predictions.
For one Euler step,
\begin{equation}
    x_{\tau+\Delta\tau}^{m}
    =
    x_{\tau}^{m}
    +
    \Delta\tau\cdot
    v_\theta^m(x_\tau^m,\tau;H_{t,\tau}),
\end{equation}
if $H_{t,\tau}=H_{t,\tau}^{\ast}+\epsilon$ (with a small disturbance $\epsilon$), a first-order approximation indicates that prediction variance scales with $\left\|
    \frac{\partial v_\theta^m}{\partial H_{t,\tau}}
    \right\|^2
    \mathrm{Var}(\epsilon)$.

To reduce the variance, following DiG-Flow~\citep{zhang2025dig} and BeingH-0.5~\citep{beingbeyond2026beingh05}, MPG computes a reliability gate $g\in(0,1]$ and modulates the residual enhancement:
\begin{equation}
\label{eq:mpg}
    \tilde{H}_{t,\tau}
    =
    H_{t,\tau}
    +
    \lambda
    \left[
    \mathbf{W}_{\mathrm{MPG}}
    \left(
    g^{\mathrm{sg}}\odot
    \mathcal{E}_{\mathrm{obs}}(H_{t,\tau})
    \right)
    +
    \mathbf{b}_{\mathrm{MPG}}
    \right],
\end{equation}
where $g^{\mathrm{sg}}=\mathrm{stopgrad}(g)$, and $\lambda$ is the weight of the residual term.

To compute $g$, we construct noise-free action and tactile anchors.
Let $Z^{\mathrm{nf,act}}$ and $Z^{\mathrm{nf,tac}}$ denote the action and tactile token embeddings encoded at noise-free level, and we mean-pool them as
\begin{equation}
    \bar{Z}^{\mathrm{act}}
    =
    \mathrm{MeanPool}(Z^{\mathrm{nf,act}}),
    \qquad
    \bar{Z}^{\mathrm{tac}}
    =
    \mathrm{MeanPool}(Z^{\mathrm{nf,tac}}).
\end{equation}
We project observation, action, and tactile features into a shared normalized space:
\begin{equation}
    \hat{H}_{t,\tau}
    =
    \mathrm{LN}
    \left(
    \mathcal{E}_{\mathrm{obs}}(H_{t,\tau})
    \right),
    \quad
    \hat{Z}^{\mathrm{act}}
    =
    \mathrm{LN}
    \left(
    \mathcal{E}_{\mathrm{act}}(\bar{Z}^{\mathrm{act}})
    \right),
    \quad
    \hat{Z}^{\mathrm{tac}}
    =
    \mathrm{LN}
    \left(
    \mathcal{E}_{\mathrm{tac}}(\bar{Z}^{\mathrm{tac}})
    \right),
\end{equation}
and quantify the feature–action-tactile distributional discrepancy in a shared, scale-invariant space using the sliced Wasserstein distance (SWD)~\citep{bonneel2015sliced,kolouri2019generalized}:
\begin{equation}
\begin{aligned}
D_{\mathrm{act}}
&=
\frac{1}{M}
\sum_{i=1}^{M}
\left\|
\mathrm{sort}\left(\theta_i^\top \hat{H}_{t,\tau}\right)
-
\mathrm{sort}\left(\theta_i^\top \hat{Z}^{\mathrm{act}}\right)
\right\|_2^2,
\\
D_{\mathrm{tac}}
&=
\frac{1}{M}
\sum_{i=1}^{M}
\left\|
\mathrm{sort}\left(\theta_i^\top \hat{H}_{t,\tau}\right)
-
\mathrm{sort}\left(\theta_i^\top \hat{Z}^{\mathrm{tac}}\right)
\right\|_2^2,
\end{aligned}
\end{equation}
where each $\theta_i$ is a random unit projection direction.
The joint discrepancy and gate are computed as
\begin{equation}
    D
    =
    \frac{1}{2}
    \left(
    D_{\mathrm{act}}+D_{\mathrm{tac}}
    \right),
    \qquad
    g
    =
    \exp(-D/\tau_g).
\end{equation}
This dual-anchor design enhances $H_{t,\tau}$ only when it aligns with both action and tactile manifolds, ensuring that the feature-dependent correction becomes increasingly insensitive when the context is unreliable (small $g$), and improving robustness under context shifts.

\section{Experiments: Towards Human-to-Robot Transfer}
\label{sec:experiments}

In the experiments, we aim to answer the following questions: (1) After tactile-based pre-training, can TTP generate the hand motion and tactile sensings well, with capabilities of generalization? (2) Even though with more training cost and burden from the newly-added tactile modality, can TTP maintain comparable performances in simulation benchmarks, and even better generalization abilities? (3) With tactile information enhanced, can TTP demonstrate great capabilities in real-world tactile-relevant tasks?

\subsection{Visualizations of Tactile-Aided Pre-Training}

% \begin{figure}[t]
%     \centering
%     \begin{tikzpicture}
%     \draw[thick] (0,0) rectangle (13,4);
%     \end{tikzpicture}
%     \caption{Visualization showcase.}
%     \label{fig:visualization}
% \end{figure}

To answer the first question, we visualize the motion generation and tactile prediction results in Figure~\ref{fig:visualization}.
We visualize the predicted hand motion and tactile sensings together, with tactile predictions rendered on the surface of MANO motions with a heatmap. 
We test the motion generation and tactile prediction on both original validation sets and those with inpaintings, including human hand inpainted with empty as well as with robot arms.
%From the visualization results, TTP can precisely predict future hand motions as well as tactile sensings in all settings, demonstrating the massive capabilities of generalization after tactile-based pre-training.

\subsection{Simulation Environment Experiments}

\definecolor{BlockA}{RGB}{180,220,255}   % 示例蓝色，请替换为你的实际颜色
\definecolor{BlockB}{RGB}{255,200,150}   % 示例橙色，请替换为你的实际颜色

\begin{table}[t]
\centering
\caption{\textbf{Success rates (\%) on LIBERO, LIBERO-plus, and RoboCasa benchmarks.}
We compare TTP with state-of-the-art models. Best results per benchmark are \textbf{bolded}, second-best are \underline{underlined}.
LIBERO: mean over 50 episodes per task. LIBERO-plus: zero-shot, mean over 70 trials per category. RoboCasa: mean over 50 trials per task (24 tasks).}
\label{tab:mixed-v2}
\footnotesize
\setlength{\tabcolsep}{3pt}
\resizebox{\textwidth}{!}{%
\begin{tabular}{l*{5}{c}*{8}{c}*{4}{c}}
\toprule
\textbf{Method} & \multicolumn{5}{c}{\textbf{LIBERO}} & \multicolumn{8}{c}{\textbf{LIBERO-plus}} & \multicolumn{4}{c}{\textbf{RoboCasa}} \\
\cmidrule(lr){2-6} \cmidrule(lr){7-14} \cmidrule(lr){15-18}
 & Spat. & Obj. & Goal & Long & Avg. & Cam. & Rob. & Lang. & Light & Bg. & Noise & Lay. & Avg. & P\&P & Door/Draw. & Others & Avg. \\
\midrule

OpenVLA \citep{kim2024openvla}
 & 84.7 & 88.4 & 79.2 & 53.7 & 76.5
 & 0.8 & 3.5 & 23.0 & 8.1 & 34.8 & 15.2 & 28.5 & 15.6
 & - & - & - & - \\
OpenVLA-OFT \citep{kim2025openvla-oft}
 & 97.6 & 98.4 & 97.9 & 94.5 & 97.1
 & \underline{56.4} & 31.9 & 79.5 & 88.7 & \underline{93.3} & 75.8 & 74.2 & 69.6
 & - & - & - & - \\

 OpenVLA-OFT\_w \citep{kim2025openvla-oft}
 & 96.2 & 98.3 & 96.2 & 90.7 & 95.3 & 10.4 & 38.7 & 70.5 & 76.8 & \textbf{93.6} & 49.9 & 69.9 & 55.8 & - & - & - & - \\
OpenVLA-OFT\_m \citep{kim2025openvla-oft}
 & 95.2 & 94.2 & 95.2 & 93.2 & 94.5 & 55.6 & 21.7 & 81.0 & 92.7 & 91.0 & 78.6 & 68.7 & 67.9 & - & - & - & - \\
NORA \citep{hung2025nora}
 & 92.2 & 95.4 & 89.4 & 74.6 & 87.9 & 2.2 & 37.0 & 65.1 & 45.7 & 58.6 & 12.8 & 62.1 & 39.0 & - & - & - & - \\
WorldVLA \citep{cen2025worldvla}
 & 85.1 & 90.9 & 84.0 & 52.4 & 78.1 & 0.1 & 27.9 & 41.6 & 43.7 & 17.1 & 10.9 & 38.0 & 25.0 & - & - & - & - \\
UniVLA \citep{bu2025univla}
 & 96.5 & 96.8 & 95.6 & 92.0 & 95.2 & 1.8 & 46.2 & 69.6 & 69.0 & 81.0 & 21.2 & 31.9 & 42.9 & - & - & - & - \\
RIPT-VLA \citep{tan2025interactive}
 & 92.7 & 95.6 & 98.4 & 87.5 & 93.6 & 55.2 & 31.2 & 77.6 & 88.4 & 91.6 & 73.5 & 74.2 & 68.4 & - & - & - & - \\

\midrule
GR00T-N1 \citep{nvidia2025gr00t}
 & 94.4 & 97.6 & 93.0 & 90.6 & 93.9
 & - & - & - & - & - & - & - & - 
 & 18.6 & 50.2 & 39.1 & 36.0 \\

\midrule
Diffusion Policy \citep{chi2025diffusion}
 & 78.5 & 87.5 & 73.5 & 64.8 & 76.1 & - & - & - & - & - & - & - & - & - & - & - & - \\
SpatialVLA \citep{qu2025spatialvla}
 & 88.2 & 89.9 & 78.6 & 55.5 & 78.1 & - & - & - & - & - & - & - & - & - & - & - & - \\
CoT-VLA \citep{zhao2025cotvla}
 & 87.5 & 91.6 & 87.6 & 69.0 & 83.9 & - & - & - & - & - & - & - & - & - & - & - & - \\
F1 \citep{lv2025f1}
 & \underline{98.2} & 97.8 & 95.4 & 91.3 & 95.7 & - & - & - & - & - & - & - & - & - & - & - & - \\
InternVLA-M1 \citep{chen2025internvla-m1}
 & 98.0 & \textbf{99.0} & 93.8 & 92.6 & 95.9 & - & - & - & - & - & - & - & - & - & - & - & - \\
Discrete Diffusion VLA \citep{liang2025discrete}
 & 97.2 & \underline{98.6} & 97.4 & 92.0 & 96.3 & - & - & - & - & - & - & - & - & - & - & - & - \\

\midrule

3DA (3D) \cite{3d_diffuser_actor}
 & - & - & - & - & - & - & - & - & - & - & - & - & - & 0.0 & 2.3 & 13.1 & 5.5 \\
DP3 (3D) \cite{Ze2024DP3}
 & - & - & - & - & - & - & - & - & - & - & - & - & - & 1.5 & 41.7 & 32.0 & 22.8 \\
GWM (3D) \cite{lu2025gwm}
 & - & - & - & - & - & - & - & - & - & - & - & - & - & 14.8 & 54.3 & 49.8 & 39.3 \\
BC (RGB 256) \cite{nasiriany2024robocasa}
 & - & - & - & - & - & - & - & - & - & - & - & - & - & 4.3 & 47.0 & 42.2 & 28.9 \\
\midrule

$\pi_0$-Fast \citep{pertsch2025fast}
 & 96.4 & 96.8 & 88.6 & 60.2 & 85.5
 & \textbf{65.1} & 21.6 & 61.0 & 73.2 & 73.2 & 74.4 & 68.8 & 61.6
 & 9.5 & 53.0 & 32.2 & 29.8 \\

$\pi_0$ \citep{black2024pi_0}
 & 98.0 & 96.8 & 94.4 & 88.4 & 94.4
 & 13.8 & 6.0 & 58.8 & 85.0 & 81.4 & 79.0 & 68.9 & 53.6
 & 14.0 & 53.1 & \textbf{58.5} & 42.4 \\

$\pi_{0.5}$ \citep{intelligence2025pi05}
 & \textbf{98.8} & 98.2 & 98.0 & 92.4 & 96.8
 & 53.0 & \textbf{50.3} & 65.7 & 83.1 & 77.3 & 53.2 & 72.7 & 65.0
 & 21.5 & 57.8 & 44.9 & 41.4 \\

{BeingH-0.5} \citep{beingbeyond2026beingh05}
 & \textbf{98.8} & 97.6 & \textbf{98.8} & \underline{96.6} & \underline{98.0}
 & 46.8 & 41.2 & 81.8 & 87.0 & 86.8 & \textbf{82.2} & 76.0 & 71.7
 & \textbf{36} & 71.7 & 57.6 & \underline{53.9} \\

\midrule

\rowcolor{BlockA!20} TTP w/o pre-training
 & 98.0 & 97.4 & 97.6 & \underline{96.6} & 97.4
 & 48.6 & 43.9 & \underline{83.2} & \underline{93.9} & 85.4 & 76.8 & \underline{82.1} & \underline{73.4}
 & 33 & \underline{72.3} & 55.6 & 52.3 \\

\rowcolor{BlockA!20} \textbf{TTP (ours)}
 & \textbf{98.8} & 98.2 & \underline{98.2} & \textbf{97.0} & \textbf{98.1}
 & 48.9 & \underline{49.5} & \textbf{84.3} & \textbf{94.6} & 87.1 & \underline{81.8} & \textbf{83.6} & \textbf{75.7}
 & \underline{35} & \textbf{76.3} & \underline{58.4} & \textbf{55.1} \\
 
\bottomrule
\end{tabular}
}
\end{table}

For the second question, we evaluate TTP on simulation benchmarks including LIBERO~\citep{liu2023libero}, LIBERO-plus~\citep{fei2025libero-plus}, and Robocasa~\citep{nasiriany2024robocasa}.
Since these benchmarks do not have tactile modalities inherently, to keep the dual-level optimization objective (action generation and tactile prediction), we use the difference between last action and current proprioceptive state as ``tactile proxy'' during post-training. Specifically, we define $o_t^{\mathrm{proxy}} = \mathrm{padding}(s_t - a_{t-1})$ as the substitution of tactile observations, with zero-padding due to mismatch between $D_{\mathrm{act}}=200$ and $D_{\mathrm{tac}}=351$.

The results on the three benchmarks are listed in Table~\ref{tab:mixed-v2}. Even though with much training cost and burden due to newly-added tactile modality with more tokens in sequence modeling, TTP still achieves comparable results against baselines in each test suite, especially with a high performance in the \texttt{LIBERO-long} suite. 
For the zero-shot results on LIBERO-plus benchmark, our policy achieves better performance in \texttt{Language}, \texttt{Light}, and \texttt{Layout}, demonstrating better overall generalization capabilities.
As for the Robocasa benchmark, TTP achieves comparable or better performances in all kinds of tasks, including \texttt{Pick \& Place}, \texttt{Doors/Drawers}, and \texttt{Others}, with relatively the best performance overall.

\subsection{Real Robot Experiments}

Experiments on real robot answer the third question. Our real robot experiments cover various platforms and embodiments, with robot arms ranging from single to dual arm, including both 6-DoF Realman and 7-DoF Franka arms. The end effectors vary from 6-DoF Inspire hands with piezoresistive tactility, to 12-DoF DexBotic hands, and even grippers with visuo-tactile sensings.

As for task settings, our tasks range from fine-grained to contact-rich settings, with various arms and hands of different degrees of freedom (DoF) and tactile sensors. From the perspective of types and contents, our tasks vary from fine-grained peeling of white radish skins, tasks with vision defects (e.g., hand-object visual occlusion during plug in), to contact-rich picking and placing fragile potato chips.

\subsubsection{Hardware Settings}

\begin{figure}[t]
    \centering
    \includegraphics[width=0.75\textwidth]{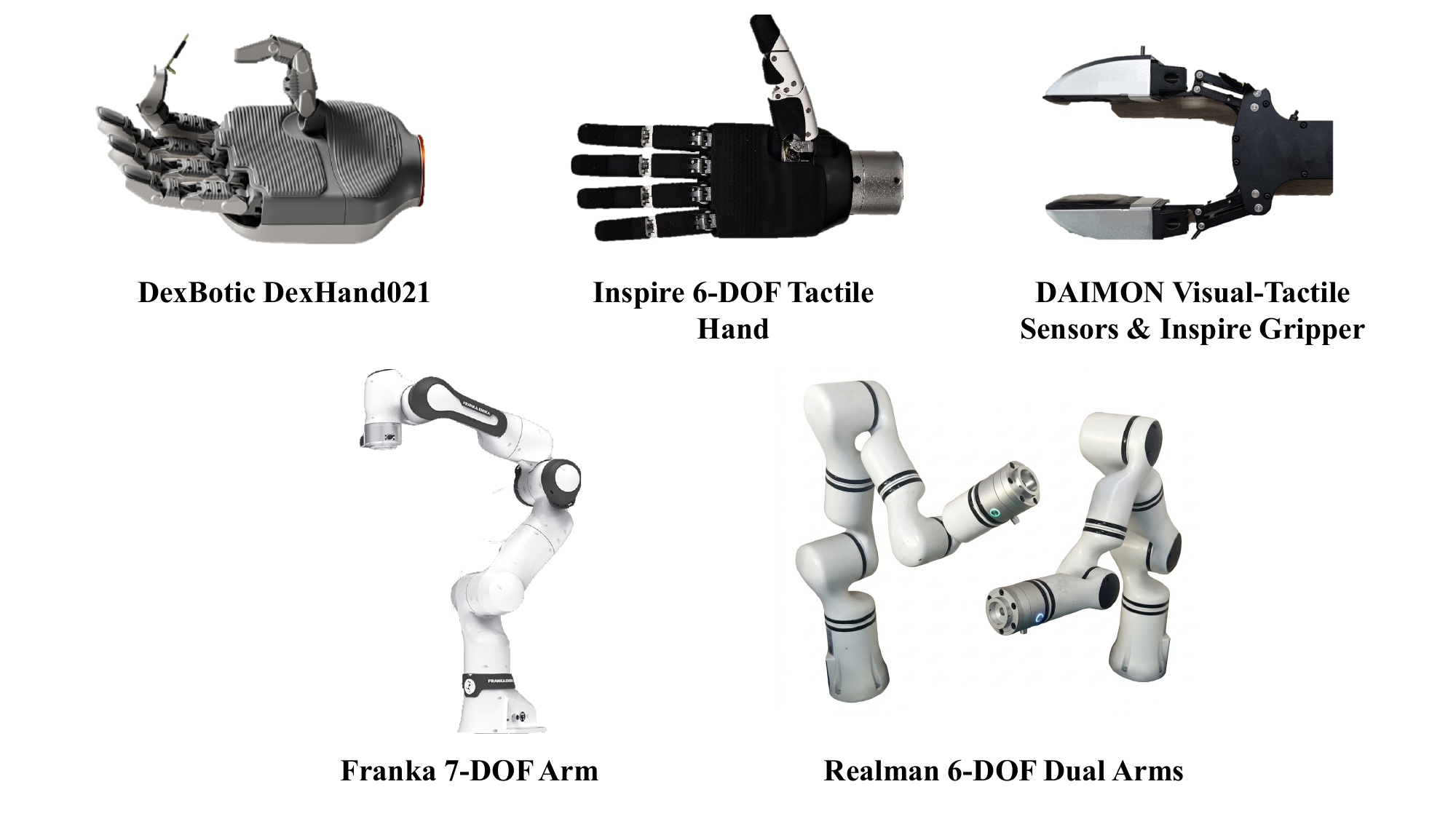}
    \caption{Hardware settings in our real-robot experiments.}
    \label{fig:hardware}
\end{figure}

In our work, we use various platforms and embodiments for real-robot experiments, as shown in Figure~\ref{fig:hardware}. Specifically, the embodiments used in our experiments are:
\begin{itemize}
    \item \textbf{Franka arm:} a 7-DoF robot arm with, with its base fixed on the tabletop. 
    \item \textbf{Realman arm(s):} two 6-DoF robot arms, with the bases fixed on the tabletop. 
    \item \textbf{Inspire hand:} a 6-DoF dexterous hand with piezoresistive tactile sensings, distributed on both the fingers and the palm.
    \item \textbf{DM-Tac gripper:} a parallel gripper with visuo-tactile sensors.
    \item \textbf{DexBotic hand:} a 12-DoF dexterous hand with 3D tactile sensings, distributed on the fingertips.
\end{itemize}

\begin{table}[t]
    \caption{Task settings and definitions. Our tasks range from fine-grained to contact rich settings, with various arms and hands of different DoFs and tactile sensors.}
    \label{tab:task-definitions}
    \centering
    \resizebox{.9\linewidth}{!}{
    \begin{tabular}{llcc}
        \toprule
        \textbf{Category} & \textbf{Task} & \textbf{Platform} & \textbf{Embodiment} \\
        \midrule
        \multirow{4}{*}{Fine-Grained} 
            & Peeling (Inspire) & Franka + Inspire & Single arm \& hand \\
            & VaseWiping (single hand) & Franka + Inspire & Single arm \& hand \\
            & VaseWiping (bimanual) & Realman + Inspire & Dual arm \& hand \\
            & Peeling (Gripper) & Franka + DM-Tac & Single arm \& gripper \\
        \midrule
        \multirow{2}{*}{Contact-Rich \& Fragile} 
            & PickPlaceChips & Franka + Inspire & Single arm \& hand \\
            & PaperFolding & Realman + Inspire & Dual arm \& hand \\
        \midrule
        \multirow{3}{*}{Vision Defect} 
            & SoftHard & Realman + Inspire & Single arm \& hand \\
            & PlugIn (Gripper) & Franka + DM-Tac & Single arm \& gripper \\
            & PlugIn (DexBotic) & Franka + DexBotic & Single arm \& hand \\
        \bottomrule
    \end{tabular}
    }
\end{table}

\subsubsection{Task Definitions}

In our real robot experiments, we have 9 tasks as listed in Table~\ref{tab:task-definitions}, categorized in different types. Table~\ref{tab:task-definitions} also lists the platforms and embodiments each task uses, including different robot arms (Franka, Realman), different end effectors (Inspire hands with piezoresistive tactile sensings, DM-Tac grippers with visuo-tactile sensings, DexBotic hands with 3D tactile sensings).

Specifically, the 9 tasks in our work are defined as follows, including task settings, embodiment configurations, and evaluation metrics:
\begin{enumerate}
    \item \textbf{Peeling (Inspire).} A single Franka arm with an Inspire hand, which grasps a peeler in hand, peels on the surface of the radish. We evaluate the performance of policies by the average length of peeled skins (centimeters). 
    \item \textbf{VaseWiping (single hand).} A single Franka arm with an Inspire hand, which grasps a sponge in hand, wipes on the surface of a fixed vase to clean the handwriting marks on it. The results are evaluated by success rate, i.e., whether the policy can successfully clean the marks.
    \item \textbf{VaseWiping (bimanual).} Two Realman arms with Inspire hands, in which the left hand grasping the vase, while the right hand grasping the sponge, cleans the handwriting marks by wiping on the surface. The results are also evaluated by success rate.
    \item \textbf{Peeling (Gripper).} A single Franka arm, with parallel grippers and DM-Tac visuo-tactile sensors, peels on the surface of the radish. We evaluate the performance of policies by the average length of peeled skins (centimeters). 
    \item \textbf{PickPlaceChips.} A single Franka arm with an Inspire hand picks up a fragile potato chip, and place it onto a plate. The success is determined only if the chips is intactly placed on the plate, without any damaging or cracking.
    \item \textbf{PaperFolding.} We use two Realman arms with Inspire hands. The left hand grasps the paper in hand, and the right hand needs to fold the paper by pinching and applying moderate pressure. If the pressure force is too small, it cannot make a proper crease; if the pressure is too large, the paper might be broken due to tearing. We apply colorful paint in the shape of a line in one of the inner surfaces of the paper in advance, and when the paper is folded, the paint will stick to the other inner surface. We measure the length of that paint line appeared on the surface, and calculate the proportion of that of the original length (manually applied paint), to serve as the evaluation metric.
    \item \textbf{SoftHard.} One single Realman arm with and Inspire hand picks up a small object in hand, and places it into one of the two boxes (which stands for ``soft'' and ``hard'' ones) according to the softness of the object. We evaluate the performance of policies by success rate, which requires the object to be placed in the correct box. 
    \item \textbf{PlugIn (Gripper).} A single Franka arm, with parallel grippers and DM-Tac visuo-tactile sensors, inserts a plug into the socket. We evaluate the performance of policies by the success rate.
    \item \textbf{PlugIn (DexBotic).} A single Franka arm, with a DexBotic hand, inserts a plug into the socket. We evaluate the performance of policies by the success rate.
\end{enumerate}

\begin{table}[t]
    \caption{Results on real-robot experiments, categorized by task types. Average task progress rates are calculated over in-distribution (ID) and out-of-distribution (OOD) tests, each for 15 trials (10 for ID and 5 for OOD).}
    \label{tab:real-robot-main-text}
    \centering
    % \resizebox{.99\linewidth}{!}{
    \begin{tabular}{lccccc}
        \toprule
        \textbf{Task Category} & \textbf{$\pi_{0.5}$} & \textbf{$\pi_{0.5}$ + tactile} & \textbf{BeingH-0.5} & \textbf{TTP w/o pre-train} & \textbf{TTP (ours)} \\
        \midrule
        Fine-grained & 43.2\% & 48.3\% & 57.3\% & 71.0\% & 96.7\% \\
        Contact-rich \& Fragile & 3.3\% & 8.0\% & 9.2\% & 49.7\% & 79.2\% \\
        Vision Defect & 17.8\% & 17.8\% & 15.6\% & 26.7\% & 37.8\% \\
        \bottomrule
    \end{tabular}
    % }
\end{table}

\begin{figure}[t]
    \centering
    \includegraphics[width=1.0\textwidth]{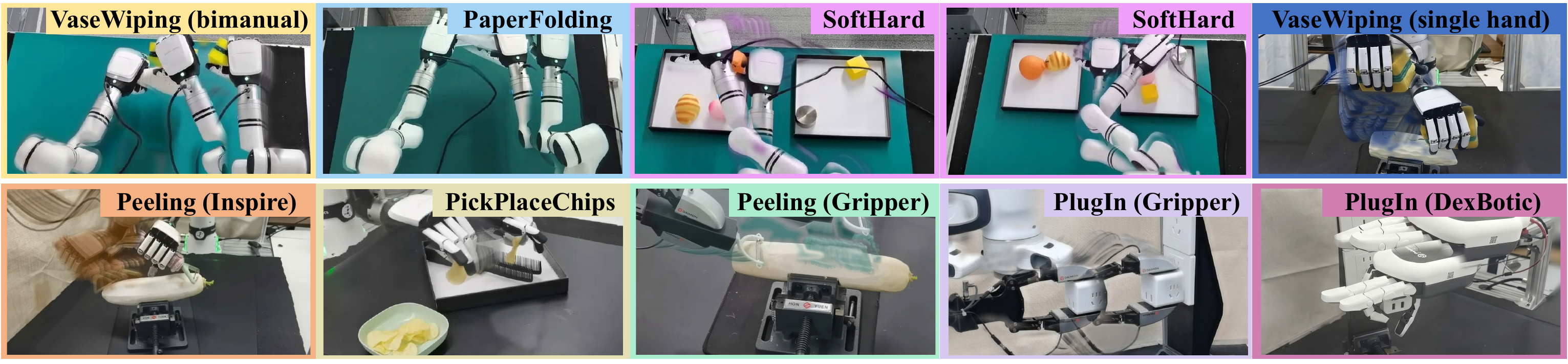}
    \caption{Real robot showcases. Our TTP demonstrate strong capabilities of precise and fine-grained manipulation, outperforming various baselines.}
    \label{fig:real-robot-showcase}
\end{figure}

\subsubsection{Results and Analysis}

Different categories of tasks are evaluated with different metrics, including (a) \texttt{Peeling}: average length of successfully peeled skins (cm), (b) \texttt{PaperFolding}: average proportion (\%) of folded length over the length of applied paint, and (c) \texttt{Others}: success rate (\%).
We calculate the average task progress rate over each task category (metrics of different dimensions using TTP (ours) as 100\% for proportional conversion), with results listed in Table~\ref{tab:real-robot-main-text}, and showcases shown in Figure~\ref{fig:real-robot-showcase}. TTP outperforms tactile-free state-of-the-art (SOTA) baselines (including BeingH-0.5~\citep{beingbeyond2026beingh05} and $\pi_{0.5}$~\citep{intelligence2025pi05}) by a large margin, demonstrating the effectiveness of tactile modality.

TTP exhibits a robust and moderate behavior mode. For instance, TTP can continuously peel the radish skin for a length of over 20 cm, while baseline methods peel only for a short length with jitter. When picking up the fragile potato chips, TTP applies a moderate grasping force, neither so strong as to crush them, nor so weak as to let them slip. In contrast, baseline methods often suffer from these problems.

Additionally, we also compare TTP against a variant without tactile-based pre-training (denoted as TTP w/o pre-training). The results that TTP achieves higher performances than TTP w/o pre-training demonstrate that through tactile-based pre-training, TTP leverages such prior knowledge, which is beneficial for human-to-robot skill transfer.

\subsubsection{In-Distribution and Out-of-Distribution Results on Real Robot Experiments}

\begin{figure}[htbp]
    \centering
    \includegraphics[width=0.9\textwidth]{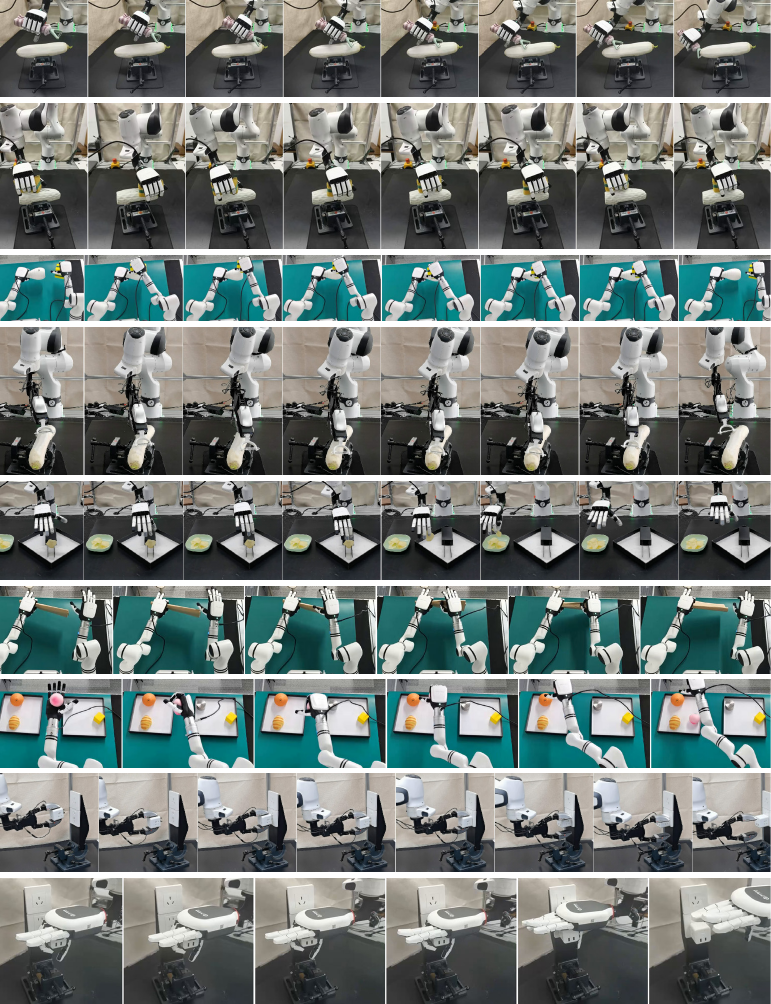}
    \caption{Demonstration showcases in our real-robot experiments (in distribution). From top to bottom are our 9 tasks: \texttt{Peeling (Inspire)}, \texttt{VaseWiping (single hand)}, \texttt{VaseWiping (bimanual)}, \texttt{Peeling (Gripper)}, \texttt{PickPlaceChips}, \texttt{PaperFolding}, \texttt{SoftHard}, \texttt{PlugIn (Gripper)}, and \texttt{PlugIn (DexBotic)}.}
    \label{fig:additional-results-ID}
\end{figure}

For our real-robot experiments, we test each methods for 10 in-distribution (ID) trails which are consistent with the post-training datasets, and calculate the average results in the corresponding metrics, with the results listed in Table~\ref{tab:additional-results-ID}. More demonstration showcases of test rollouts are shown in Figure~\ref{fig:additional-results-ID}.

\begin{table}[t]
    \caption{Detailed results on real-robot experiments (in distribution), averaged over 10 test trails. Metrics including: (a) Peeling: average length of successfully peeled skins (cm), (b) PaperFolding: average proportion (\%) of folded length over the length of applied paint, and (c) Others: success rate (\%).}
    \label{tab:additional-results-ID}
    \centering
    \resizebox{.99\linewidth}{!}{
    \begin{tabular}{llccccc}
        \toprule
        \textbf{Category} & \textbf{Task} & \textbf{$\pi_{0.5}$} & \textbf{$\pi_{0.5}$ + tactile} & \textbf{BeingH-0.5} & \textbf{TTP w/o pre-train} & \textbf{TTP (ours)} \\
        \midrule
        \multirow{4}{*}{Fine-Grained} 
            & Peeling (Inspire) & 10.63 cm & 9.27 cm & 12.49 cm & 14.65 cm & 23.33 cm \\
            & VaseWiping (single hand) & 30\% & 50\% & 50\% & 70\% & 100\% \\
            & VaseWiping (bimanual) & 50\% & 40\% & 70\% & 60\% & 90\% \\
            & Peeling (Gripper) & 10.39 cm & 12.02 cm & 11.37 cm & 14.48 cm & 15.24 cm \\
        \midrule
        \multirow{2}{*}{Contact-Rich \& Fragile} 
            & PickPlaceChips & 10\% & 20\% & 10\% & 60\% & 80\% \\
            & PaperFolding & 0\% & 4\% & 12\% & 57\% & 84\% \\
        \midrule
        \multirow{3}{*}{Vision Defect} 
            & SoftHard & 50\% & 60\% & 40\% & 80\% & 80\% \\
            & PlugIn (Gripper) & 0\% & 0\% & 0\% & 0\% & 20\% \\
            & PlugIn (DexBotic) & 0\% & 0\% & 0\% & 10\% & 10\% \\
        \bottomrule
    \end{tabular}
    }
\end{table}

\begin{table}[t]
    \caption{Detailed results on real-robot experiments (out of distribution), averaged over 5 test trails. Metrics including: (a) Peeling: average length of successfully peeled skins (cm), (b) PaperFolding: average proportion (\%) of folded length over the length of applied paint, and (c) Others: success rate (\%).}
    \label{tab:additional-results-OOD}
    \centering
    \resizebox{.99\linewidth}{!}{
    \begin{tabular}{llccccc}
        \toprule
        \textbf{Category} & \textbf{Task} & \textbf{$\pi_{0.5}$} & \textbf{$\pi_{0.5}$ + tactile} & \textbf{BeingH-0.5} & \textbf{TTP w/o pre-train} & \textbf{TTP (ours)} \\
        \midrule
        \multirow{4}{*}{Fine-Grained} 
            & Peeling (Inspire) & 5.74cm & 5.48cm & 9.28 cm & 11.25cm & 19.12 cm \\
            & VaseWiping (single hand) & 20\% & 20\% & 40\% & 80\% & 100\% \\
            & VaseWiping (bimanual) & 40\% & 60\% & 60\% & 40\% & 80\% \\
            & Peeling (Gripper) & 6.68 cm & 8.71 cm & 7.04 cm & 15.83 cm & 16.24 cm \\
        \midrule
        \multirow{2}{*}{Contact-Rich \& Fragile} 
            & PickPlaceChips & 0\% & 0\% & 0\% & 40\% & 60\% \\
            & PaperFolding & 0\% & 0\% & 11\% & 24\% & 87\% \\
        \midrule
        \multirow{3}{*}{Vision Defect} 
            & SoftHard & 60\% & 40\% & 60\% & 60\% & 80\% \\
            & PlugIn (Gripper) & 0\% & 0\% & 0\% & 0\% & 20\% \\
            & PlugIn (DexBotic) & 0\% & 0\% & 0\% & 0\% & 20\% \\
        \bottomrule
    \end{tabular}
    }
\end{table}

\begin{figure}[htbp]
    \centering
    \includegraphics[width=0.89\textwidth]{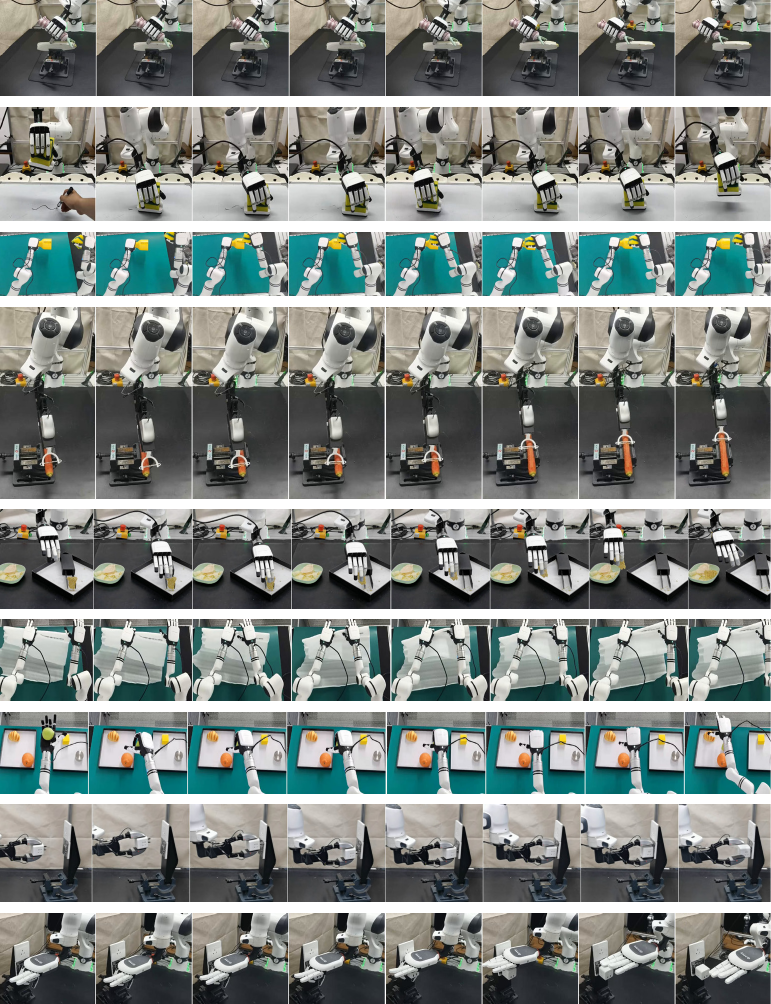}
    \caption{Demonstration showcases in our real-robot experiments (out of distribution). 
    % From top to bottom are our 9 tasks: \texttt{Peeling (Inspire)}, \texttt{VaseWiping (single hand)}, \texttt{VaseWiping (bimanual)}, \texttt{Peeling (Gripper)}, \texttt{PickPlaceChips}, \texttt{PaperFolding}, \texttt{SoftHard}, \texttt{PlugIn (Gripper)}, and \texttt{PlugIn (DexBotic)}. 
    For \texttt{Peeling} tasks, we demonstrate object generalization including carrots and cucumbers. For \texttt{VaseWiping} tasks, we demonstrate object generalization including unseen vases and even wiping a whiteboard. For \texttt{PickPlaceChips}, we demonstrate location and object generalization with crispy instant noodles. For \texttt{PaperFolding}, we demonstrate object generalization with various unseen paper-like objects (A4, cardboards, and a piece of soft cloth). For \texttt{SoftHard}, we demonstrate object generalization including unseen soft and hard objects. For \texttt{PlugIn} tasks, we demonstrate visual interruption by painting the socket into black.
    }
    \label{fig:additional-results-OOD}
\end{figure}

In addition, we test each methods for 5 out-of-distribution (OOD) trails, and calculate the average results in the corresponding metrics, with the results listed in Table~\ref{tab:additional-results-OOD}. More demonstration showcases of test rollouts are shown in Figure~\ref{fig:additional-results-OOD}. The OOD generalization categories include object generalization (peeling carrots and cucumbers, folding papers with different materials, wiping new vases, distinguishing unseen soft and hard objects), location generalization (potato chips located at different locations), scene generalization (plugging in a socket in black), etc.

\subsection{Ablation Study}

\subsubsection{Ablation Studies on Model Design}

In this part, we provide ablations on whether using Tactile-Action MPG and whether using tactile expert for future tactile prediction will help the model training. We conduct such an ablation study during the pre-training phase (150k training steps each), testing motion prediction performances under each different settings on the validation set. Specifically, we use MPJPE, PA-MPJPE, MPJAE, PA-MPJAE as metrics. We use \texttt{w/o MPG} to denote our training methods without Tactile-Action MPG, and \texttt{w/o tac-pred} to denote our methods without future tactile prediction. 

The test results on validation set are shown in Table~\ref{tab:ablation-main-text}. The results show that both excluding MPG and excluding tactile prediction will have a negative effect on training performances, yielding higher motion prediction errors. Such results demonstrate the effectiveness of our design on Tactile-Action MPG and tactile expert modules.

\begin{table}[t]
\centering
\caption{Ablations on whether using Tactile-Action MPG and whether using tactile expert for future tactile prediction will help the model training. We show the motion prediction errors on validation sets during tactile pre-training.}
\label{tab:ablation-main-text}
\resizebox{0.73\textwidth}{!}{%
\begin{tabular}{lcccc}
\toprule
\textbf{Method} & \textbf{MPJPE} & \textbf{PA-MPJPE} & \textbf{MPJAE} & \textbf{PA-MPJAE} \\
\midrule
TTP w/o MPG w/o tac-pred & 25.5850 & 0.8622 & 0.0277 & 0.0620 \\
TTP w/o MPG & 24.7597 & 0.8151 & 0.0267 & 0.0598 \\
TTP w/o tac-pred & 24.5518 & 0.8009 & 0.0263 & 0.0583 \\
TTP (ours) & \textbf{23.5711} & \textbf{0.7877} & \textbf{0.0257} & \textbf{0.0559} \\
\bottomrule
\end{tabular}
}
\end{table}

\subsubsection{Ablation Studies on Scaling}

We also conduct ablation studies on scaling up in dataset size during pre-training, focusing on how data amount affects training performances.
We conduct such an ablation study during the pre-training phase (150k training steps each, with different amount of training data including 10\%, 25\%, 50\%, 75\%, and 100\% uniformly sampled from the original pre-training dataset), testing motion prediction performances under each different settings on the fixed validation set. Similarly, we use MPJPE, PA-MPJPE, MPJAE, PA-MPJAE as metrics.

The test results on validation set are shown in Table~\ref{tab:ablation-scaling}. The results show that the motion prediction error decreases as more training data are used, which demonstrate that our proposed tactile-based pre-training can scale up.

\begin{table}[t]
\centering
\caption{Ablations on the relationships between training data amount (percentage uniformly sampled from the original pre-training dataset) and model performance during tactile pre-training.}
\label{tab:ablation-scaling}
\resizebox{0.73\textwidth}{!}{%
\begin{tabular}{ccccc}
\toprule
\textbf{Percentage of Training Data} & \textbf{MPJPE} & \textbf{PA-MPJPE} & \textbf{MPJAE} & \textbf{PA-MPJAE} \\
\midrule
10\% & 33.1917  & 1.4066  & 0.0421  & 0.0958  \\
25\% & 29.6462  & 1.2563  & 0.0374  & 0.0806  \\
50\% & 25.4919  & 1.1336  & 0.0335  & 0.0698  \\
75\% & 24.4753  & 0.9162  & 0.0295  & 0.0623  \\
100\% (ours) & \textbf{23.5711} & \textbf{0.7877} & \textbf{0.0257} & \textbf{0.0559} \\
\bottomrule
\end{tabular}
}
\end{table}

\section{Conclusion}

In this paper, we propose TTP, the first system with human-centric tactile pre-training with egocentric videos, language instructions, and paired action and tactile annotations. By tactile pre-training, TTP manages to align tactile sensings with other modalities implicitly, which can handle both tactile observation inputs and future tactile predictions, modeling the tactile dynamics in the environment. In tactile-relevant and contact rich tasks that need dexterous and fine-grained manipulation, TTP demonstrates excellent performances under extensive experiments, outperforming non-pre-trained and non-tactile baselines. TTP paves a pathway to scalable tactile pre-training, revealing the capabilities of human-to-robot skill transfer.

% \section{Limitations}

% Our work has some limitations that could be addressed in the future. On the one hand, some fine-grained tasks have relatively low success rate, with typical failure modes. This is largely due to the low-quality of teleoperated datasets of post-training.
% On the other hand, our model is weak in language following, since the pre-training datasets do not contain data in such aspects. Adding tactile-language-action data may help.

\bibliographystyle{plainnat}
\bibliography{ref}

@article{zhang2025dig,
  title={DiG-Flow: Discrepancy-Guided Flow Matching for Robust VLA Models},
  author={Zhang, Wanpeng and Wang, Ye and Luo, Hao and Yuan, Haoqi and Feng, Yicheng and Zheng, Sipeng and Jin, Qin and Lu, Zongqing},
  journal={arXiv preprint arXiv:2512.01715},
  year={2025}
}

@article{beingbeyond2026beingh05,
  title={Being-H0.5: Scaling Human-Centric Robot Learning for Cross-Embodiment Generalization},
  author={Luo, Hao and Wang, Ye and Zhang, Wanpeng and Zheng, Sipeng and Xi, Ziheng and Xu, Chaoyi and Xu, Haiweng and Yuan, Haoqi and Zhang, Chi and Wang, Yiqing and others},
  journal={arXiv preprint arXiv:2601.12993},
  year={2026}
}

@article{bonneel2015sliced,
  title={Sliced and radon wasserstein barycenters of measures},
  author={Bonneel, Nicolas and Rabin, Julien and Peyr{\'e}, Gabriel and Pfister, Hanspeter},
  journal={Journal of Mathematical Imaging and Vision},
  volume={51},
  number={1},
  pages={22--45},
  year={2015},
  publisher={Springer}
}

@article{kolouri2019generalized,
  title={Generalized sliced wasserstein distances},
  author={Kolouri, Soheil and Nadjahi, Kimia and Simsekli, Umut and Badeau, Roland and Rohde, Gustavo},
  journal={Advances in neural information processing systems},
  volume={32},
  year={2019}
}

@article{wang2025internvl3.5,
  title={Internvl3.5: Advancing open-source multimodal models in versatility, reasoning, and efficiency},
  author={Wang, Weiyun and Gao, Zhangwei and Gu, Lixin and Pu, Hengjun and Cui, Long and Wei, Xingguang and Liu, Zhaoyang and Jing, Linglin and Ye, Shenglong and Shao, Jie and others},
  journal={arXiv preprint arXiv:2508.18265},
  year={2025}
}

@article{luo2025beingh0,
  title={Being-h0: vision-language-action pretraining from large-scale human videos},
  author={Luo, Hao and Feng, Yicheng and Zhang, Wanpeng and Zheng, Sipeng and Wang, Ye and Yuan, Haoqi and Liu, Jiazheng and Xu, Chaoyi and Jin, Qin and Lu, Zongqing},
  journal={arXiv preprint arXiv:2507.15597},
  year={2025}
}

@article{zhang2025unitachand,
  title={UniTacHand: Unified Spatio-Tactile Representation for Human to Robotic Hand Skill Transfer},
  author={Zhang, Chi and Cai, Penglin and Yuan, Haoqi and Xu, Chaoyi and Lu, Zongqing},
  journal={arXiv preprint arXiv:2512.21233},
  year={2025}
}

@article{romero2022mano,
  title={Embodied hands: Modeling and capturing hands and bodies together},
  author={Romero, Javier and Tzionas, Dimitrios and Black, Michael J},
  journal={arXiv preprint arXiv:2201.02610},
  year={2022}
}

@article{liu2023libero,
  title={Libero: Benchmarking knowledge transfer for lifelong robot learning},
  author={Liu, Bo and Zhu, Yifeng and Gao, Chongkai and Feng, Yihao and Liu, Qiang and Zhu, Yuke and Stone, Peter},
  journal={Advances in Neural Information Processing Systems},
  volume={36},
  pages={44776--44791},
  year={2023}
}

@article{fei2025libero-plus,
    title={LIBERO-Plus: In-depth Robustness Analysis of Vision-Language-Action Models},
    author={Senyu Fei and Siyin Wang and Junhao Shi and Zihao Dai and Jikun Cai and Pengfang Qian and Li Ji and Xinzhe He and Shiduo Zhang and Zhaoye Fei and Jinlan Fu and Jingjing Gong and Xipeng Qiu},
    journal = {arXiv preprint arXiv:2510.13626},
    year={2025},
}

@article{nasiriany2024robocasa,
  title={Robocasa: Large-scale simulation of everyday tasks for generalist robots},
  author={Nasiriany, Soroush and Maddukuri, Abhiram and Zhang, Lance and Parikh, Adeet and Lo, Aaron and Joshi, Abhishek and Mandlekar, Ajay and Zhu, Yuke},
  journal={arXiv preprint arXiv:2406.02523},
  year={2024}
}

@article{chi2025diffusion,
  title={Diffusion policy: Visuomotor policy learning via action diffusion},
  author={Chi, Cheng and Xu, Zhenjia and Feng, Siyuan and Cousineau, Eric and Du, Yilun and Burchfiel, Benjamin and Tedrake, Russ and Song, Shuran},
  journal={The International Journal of Robotics Research},
  volume={44},
  number={10-11},
  pages={1684--1704},
  year={2025},
  publisher={Sage Publications Sage UK: London, England}
}

@article{kim2024openvla,
  title={Openvla: An open-source vision-language-action model},
  author={Kim, Moo Jin and Pertsch, Karl and Karamcheti, Siddharth and Xiao, Ted and Balakrishna, Ashwin and Nair, Suraj and Rafailov, Rafael and Foster, Ethan and Lam, Grace and Sanketi, Pannag and others},
  journal={arXiv preprint arXiv:2406.09246},
  year={2024}
}

@article{qu2025spatialvla,
  title={Spatialvla: Exploring spatial representations for visual-language-action model},
  author={Qu, Delin and Song, Haoming and Chen, Qizhi and Yao, Yuanqi and Ye, Xinyi and Ding, Yan and Wang, Zhigang and Gu, JiaYuan and Zhao, Bin and Wang, Dong and others},
  journal={arXiv preprint arXiv:2501.15830},
  year={2025}
}

@inproceedings{zhao2025cotvla,
  title={Cot-vla: Visual chain-of-thought reasoning for vision-language-action models},
  author={Zhao, Qingqing and Lu, Yao and Kim, Moo Jin and Fu, Zipeng and Zhang, Zhuoyang and Wu, Yecheng and Li, Zhaoshuo and Ma, Qianli and Han, Song and Finn, Chelsea and others},
  booktitle={Proceedings of the Computer Vision and Pattern Recognition Conference},
  pages={1702--1713},
  year={2025}
}

@article{pertsch2025fast,
  title={Fast: Efficient action tokenization for vision-language-action models},
  author={Pertsch, Karl and Stachowicz, Kyle and Ichter, Brian and Driess, Danny and Nair, Suraj and Vuong, Quan and Mees, Oier and Finn, Chelsea and Levine, Sergey},
  journal={arXiv preprint arXiv:2501.09747},
  year={2025}
}

@article{nvidia2025gr00t,
  title={Gr00t n1: An open foundation model for generalist humanoid robots},
  author={Nvidia, J Bjorck and Castaneda, Fernando and Cherniadev, N and Da, X and Ding, R and Fan, L and Fang, Y and Fox, D and Hu, F and Huang, S and others},
  journal={arXiv preprint arXiv:2503.14734},
  year={2025}
}

@article{black2024pi_0,
  title={{$\pi_{0}$}: A Vision-Language-Action Flow Model for General Robot Control},
  author={Black, Kevin and Brown, Noah and Driess, Danny and Esmail, Adnan and Equi, Michael and Finn, Chelsea and Fusai, Niccolo and Groom, Lachy and Hausman, Karol and Ichter, Brian and others},
  journal={arXiv preprint arXiv:2410.24164},
  year={2024}
}

@article{lv2025f1,
  title={F1: A vision-language-action model bridging understanding and generation to actions},
  author={Lv, Qi and Kong, Weijie and Li, Hao and Zeng, Jia and Qiu, Zherui and Qu, Delin and Song, Haoming and Chen, Qizhi and Deng, Xiang and Pang, Jiangmiao},
  journal={arXiv preprint arXiv:2509.06951},
  year={2025}
}

@article{chen2025internvla-m1,
  title={Internvla-m1: A spatially guided vision-language-action framework for generalist robot policy},
  author={Chen, Xinyi and Chen, Yilun and Fu, Yanwei and Gao, Ning and Jia, Jiaya and Jin, Weiyang and Li, Hao and Mu, Yao and Pang, Jiangmiao and Qiao, Yu and others},
  journal={arXiv preprint arXiv:2510.13778},
  year={2025}
}

@article{liang2025discrete,
  title={Discrete diffusion vla: Bringing discrete diffusion to action decoding in vision-language-action policies},
  author={Liang, Zhixuan and Li, Yizhuo and Yang, Tianshuo and Wu, Chengyue and Mao, Sitong and Nian, Tian and Pei, Liuao and Zhou, Shunbo and Yang, Xiaokang and Pang, Jiangmiao and others},
  journal={arXiv preprint arXiv:2508.20072},
  year={2025}
}

@article{intelligence2025pi05,
  title={{$\pi_{0.5}$}: A Vision-Language-Action Model with Open-World Generalization},
  author={Physical Intelligence and Black, Kevin and Brown, Noah and Darpinian, James and Dhabalia, Karan and Driess, Danny and Esmail, Adnan and Equi, Michael and Finn, Chelsea and Fusai, Niccolo and others},
  journal={arXiv preprint arXiv:2504.16054},
  year={2025}
}

@article{kim2025openvla-oft,
  title={Fine-tuning vision-language-action models: Optimizing speed and success},
  author={Kim, Moo Jin and Finn, Chelsea and Liang, Percy},
  journal={arXiv preprint arXiv:2502.19645},
  year={2025}
}

@article{3d_diffuser_actor,
  author = {Ke, Tsung-Wei and Gkanatsios, Nikolaos and Fragkiadaki, Katerina},
  title = {3D Diffuser Actor: Policy Diffusion with 3D Scene Representations},
  journal = {Arxiv},
  year = {2024}
}

@inproceedings{Ze2024DP3,
	title={3D Diffusion Policy: Generalizable Visuomotor Policy Learning via Simple 3D Representations},
	author={Yanjie Ze and Gu Zhang and Kangning Zhang and Chenyuan Hu and Muhan Wang and Huazhe Xu},
	booktitle={Proceedings of Robotics: Science and Systems (RSS)},
	year={2024}
}

@inproceedings{lu2025gwm,
  title={Gwm: Towards scalable gaussian world models for robotic manipulation},
  author={Lu, Guanxing and Jia, Baoxiong and Li, Puhao and Chen, Yixin and Wang, Ziwei and Tang, Yansong and Huang, Siyuan},
  booktitle={Proceedings of the IEEE/CVF International Conference on Computer Vision},
  pages={9263--9274},
  year={2025}
}

@article{hung2025nora,
  title={Nora: A small open-sourced generalist vision language action model for embodied tasks},
  author={Hung, Chia-Yu and Sun, Qi and Hong, Pengfei and Zadeh, Amir and Li, Chuan and Tan, U and Majumder, Navonil and Poria, Soujanya and others},
  journal={arXiv preprint arXiv:2504.19854},
  year={2025}
}

@article{cen2025worldvla,
  title={Worldvla: Towards autoregressive action world model},
  author={Cen, Jun and Yu, Chaohui and Yuan, Hangjie and Jiang, Yuming and Huang, Siteng and Guo, Jiayan and Li, Xin and Song, Yibing and Luo, Hao and Wang, Fan and others},
  journal={arXiv preprint arXiv:2506.21539},
  year={2025}
}

@article{bu2025univla,
  title={Univla: Learning to act anywhere with task-centric latent actions},
  author={Bu, Qingwen and Yang, Yanting and Cai, Jisong and Gao, Shenyuan and Ren, Guanghui and Yao, Maoqing and Luo, Ping and Li, Hongyang},
  journal={arXiv preprint arXiv:2505.06111},
  year={2025}
}

@article{tan2025interactive,
  title={Interactive post-training for vision-language-action models},
  author={Tan, Shuhan and Dou, Kairan and Zhao, Yue and Kr{\"a}henb{\"u}hl, Philipp},
  journal={arXiv preprint arXiv:2505.17016},
  year={2025}
}

@article{hao2026tla,
  title={TLA: tactile-language-action model for contact-rich manipulation},
  author={Hao, Peng and Zhang, Chaofan and Li, Dingzhe and Cao, Xiaoge and Hao, Xiaoshuai and Cui, Shaowei and Wang, Shuo},
  journal={Robot Learning},
  volume={3},
  number={1},
  pages={17--18},
  year={2026},
  publisher={ELSPublishing}
}

@article{zhang2026vtla,
  title={Vtla: Vision-tactile-language-action model with preference learning for insertion manipulation},
  author={Zhang, Chaofan and Hao, Peng and Cao, Xiaoge and Hao, Xiaoshuai and Cui, Shaowei and Wang, Shuo},
  journal={Biomimetic Intelligence and Robotics},
  pages={100333},
  year={2026},
  publisher={Elsevier}
}

@article{huang2025tactile,
  title={Tactile-VLA: unlocking vision-language-action model's physical knowledge for tactile generalization},
  author={Huang, Jialei and Wang, Shuo and Lin, Fanqi and Hu, Yihang and Wen, Chuan and Gao, Yang},
  journal={arXiv preprint arXiv:2507.09160},
  year={2025}
}

@article{zhang2026craft,
  title={CRAFT: Adapting VLA Models to Contact-rich Manipulation via Force-aware Curriculum Fine-tuning},
  author={Zhang, Yike and Wang, Yaonan and Sun, Xinxin and Huang, Kaizhen and Xu, Zhiyuan and Ji, Junjie and Che, Zhengping and Tang, Jian and Sun, Jingtao},
  journal={arXiv preprint arXiv:2602.12532},
  year={2026}
}

@article{li2026favla,
  title={FAVLA: A Force-Adaptive Fast-Slow VLA model for Contact-Rich Robotic Manipulation},
  author={Li, Yao and Tang, Peiyuan and Zhang, Wuyang and Zhu, Chengyang and Duan, Yifan and Shi, Weikai and Zhang, Xiaodong and Yang, Zijiang and Ji, Jianmin and Zhang, Yanyong},
  journal={arXiv preprint arXiv:2602.23648},
  year={2026}
}

@article{li2026atvla,
  title={AT-VLA: Adaptive Tactile Injection for Enhanced Feedback Reaction in Vision-Language-Action Models},
  author={Li, Xiaoqi and Cai, Muhe and Xu, Jiadong and Zhu, Juan and Fan, Hongwei and Shen, Yan and Ren, Guangrui and Dong, Hao},
  journal={arXiv preprint arXiv:2605.07308},
  year={2026}
}

@article{cheng2025omnivtla,
  title={Omnivtla: Vision-tactile-language-action model with semantic-aligned tactile sensing},
  author={Cheng, Zhengxue and Zhang, Yiqian and Zhang, Wenkang and Li, Haoyu and Wang, Keyu and Song, Li and Zhang, Hengdi},
  journal={arXiv preprint arXiv:2508.08706},
  year={2025}
}

@article{bi2026vlatouch,
  title={VLA-Touch: Enhancing Vision-Language-Action Model with Dual-Level Tactile Feedback},
  author={Bi, Jianxin and Ma, Kevin Yuchen and Hao, Ce and Zheng, Mike Shou and Soh, Harold},
  journal={IEEE Robotics and Automation Letters},
  year={2026},
  publisher={IEEE}
}

@article{gubernatorov2026hapticvla,
  title={HapticVLA: Contact-Rich Manipulation via Vision-Language-Action Model without Inference-Time Tactile Sensing},
  author={Gubernatorov, Konstantin and Sannikov, Mikhail and Mikhalchuk, Ilya and Kuznetsov, Egor and Artemov, Makar and Ouwatobi, Ogunwoye Faith and Fernando, Marcelino and Asanov, Artem and Guo, Ziang and Tsetserukou, Dzmitry},
  journal={arXiv preprint arXiv:2603.15257},
  year={2026}
}

@inproceedings{grauman2022ego4d,
  title={Ego4d: Around the world in 3,000 hours of egocentric video},
  author={Grauman, Kristen and Westbury, Andrew and Byrne, Eugene and Chavis, Zachary and Furnari, Antonino and Girdhar, Rohit and Hamburger, Jackson and Jiang, Hao and Liu, Miao and Liu, Xingyu and others},
  booktitle={Proceedings of the IEEE/CVF conference on computer vision and pattern recognition},
  pages={18995--19012},
  year={2022}
}

@article{hoque2025egodex,
  title={EgoDex: Learning Dexterous Manipulation from Large-Scale Egocentric Video},
  author={Hoque, Ryan and Huang, Peide and Yoon, David J and Sivapurapu, Mouli and Zhang, Jian},
  journal={arXiv preprint arXiv:2505.11709},
  year={2025}
}

@article{yang2025egovla,
  title={Egovla: Learning vision-language-action models from egocentric human videos},
  author={Yang, Ruihan and Yu, Qinxi and Wu, Yecheng and Yan, Rui and Li, Borui and Cheng, An-Chieh and Zou, Xueyan and Fang, Yunhao and Cheng, Xuxin and Qiu, Ri-Zhao and others},
  journal={arXiv preprint arXiv:2507.12440},
  year={2025}
}

@article{wi2026tactalign,
  title={TactAlign: Human-to-Robot Policy Transfer via Tactile Alignment},
  author={Wi, Youngsun and Yin, Jessica and Xiang, Elvis and Sharma, Akash and Malik, Jitendra and Mukadam, Mustafa and Fazeli, Nima and Hellebrekers, Tess},
  journal={arXiv preprint arXiv:2602.13579},
  year={2026}
}

@inproceedings{chi2024universal,
  title={Universal Manipulation Interface: In-The-Wild Robot Teaching Without In-The-Wild Robots},
  author={Chi, Cheng and Xu, Zhenjia and Pan, Chuer and Cousineau, Eric and Burchfiel, Benjamin and Feng, Siyuan and Tedrake, Russ and Song, Shuran},
  year={2024},
  organization={Robotics: Science and Systems}
}

@inproceedings{bharadhwaj2025gen2act,
  title={Gen2Act: Human Video Generation in Novel Scenarios enables Generalizable Robot Manipulation},
  author={Bharadhwaj, Homanga and Dwibedi, Debidatta and Gupta, Abhinav and Tulsiani, Shubham and Doersch, Carl and Xiao, Ted and Shah, Dhruv and Xia, Fei and Sadigh, Dorsa and Kirmani, Sean},
  booktitle={Conference on Robot Learning},
  pages={3936--3951},
  year={2025},
  organization={PMLR}
}

@inproceedings{xie2026human2robot,
  title={Human2robot: Learning robot actions from paired human-robot videos},
  author={Xie, Sicheng and Cao, Haidong and Weng, Zejia and Xing, Zhen and Chen, Haoran and Shen, Shiwei and Leng, Jiaqi and Wu, Zuxuan and Jiang, Yu-Gang},
  booktitle={Proceedings of the AAAI Conference on Artificial Intelligence},
  volume={40},
  pages={11078--11086},
  year={2026}
}

@inproceedings{kim2025uniskill,
  title={UniSkill: Imitating Human Videos via Cross-Embodiment Skill Representations},
  author={Kim, Hanjung and Kang, Jaehyun and Kang, Hyolim and Cho, Meedeum and Kim, Seon Joo and Lee, Youngwoon},
  booktitle={Conference on Robot Learning},
  pages={4269--4294},
  year={2025},
  organization={PMLR}
}

@article{zhou2026traj2action,
      title={Traj2Action: A Co-Denoising Framework for Trajectory-Guided Human-to-Robot Skill Transfer}, 
      author={Han Zhou and Jinjin Cao and Liyuan Ma and Xueji Fang and Guo-jun Qi},
      journal={arXiv preprint arXiv:2510.00491},
      year={2025},
}

@article{heppert2026scaling,
  title={Scaling Single Human Demonstrations for Imitation Learning using Generative Foundational Models},
  author={Heppert, Nick and Nguyen, Minh Quang and Valada, Abhinav},
  journal={arXiv preprint arXiv:2602.12734},
  year={2026}
}

@inproceedings{kareer2025egomimic,
  title={Egomimic: Scaling imitation learning via egocentric video},
  author={Kareer, Simar and Patel, Dhruv and Punamiya, Ryan and Mathur, Pranay and Cheng, Shuo and Wang, Chen and Hoffman, Judy and Xu, Danfei},
  booktitle={2025 IEEE International Conference on Robotics and Automation (ICRA)},
  pages={13226--13233},
  year={2025},
  organization={IEEE}
}

@article{zheng2026egoscale,
  title={Egoscale: Scaling dexterous manipulation with diverse egocentric human data},
  author={Zheng, Ruijie and Niu, Dantong and Xie, Yuqi and Wang, Jing and Xu, Mengda and Jiang, Yunfan and Casta{\~n}eda, Fernando and Hu, Fengyuan and Tan, You Liang and Fu, Letian and others},
  journal={arXiv preprint arXiv:2602.16710},
  year={2026}
}

@article{zhang2026unidex,
  title={Unidex: A robot foundation suite for universal dexterous hand control from egocentric human videos},
  author={Zhang, Gu and Xu, Qicheng and Zhang, Haozhe and Ma, Jianhan and He, Long and Bao, Yiming and Ping, Zeyu and Yuan, Zhecheng and Lu, Chenhao and Yuan, Chengbo and others},
  journal={arXiv preprint arXiv:2603.22264},
  year={2026}
}

@inproceedings{liu2025vtdexmanip,
  title={VTDexmanip: A dataset and benchmark for visual-tactile pretraining and dexterous manipulation with reinforcement learning},
  author={Liu, Qingtao and Cui, Yu and Sun, Zhengnan and Li, Gaofeng and Chen, Jiming and Ye, Qi},
  booktitle={The Thirteenth International Conference on Learning Representations},
  year={2025}
}

@inproceedings{xue2025reactive,
  title     = {Reactive Diffusion Policy: Slow-Fast Visual-Tactile Policy Learning for Contact-Rich Manipulation},
  author    = {Xue, Han and Ren, Jieji and Chen, Wendi and Zhang, Gu and Fang, Yuan and Gu, Guoying and Xu, Huazhe and Lu, Cewu},
  booktitle = {Proceedings of Robotics: Science and Systems (RSS)},
  year      = {2025}
}

@article{zheng2026omnivta,
  title={OmniVTA: Visuo-tactile world modeling for contact-rich robotic manipulation},
  author={Zheng, Yuhang and Gu, Songen and Li, Weize and Zheng, Yupeng and Zang, Yujie and Tian, Shuai and Li, Xiang and Hao, Ce and Gao, Chen and Liu, Si and others},
  journal={arXiv preprint arXiv:2603.19201},
  year={2026}
}

@article{miller2026enhancing,
  title={Enhancing tactile-based reinforcement learning for robotic control},
  author={Miller, Elle and McInroe, Trevor and Abel, David and Mac Aodha, Oisin and Vijayakumar, Sethu},
  journal={Advances in Neural Information Processing Systems},
  volume={38},
  pages={129460--129494},
  year={2025}
}

@article{hu2025tactile,
  title={Tactile-based Reinforcement Learning for Adaptive Grasping under Observation Uncertainties},
  author={Hu, Xiao and Ye, Yang},
  journal={arXiv preprint arXiv:2505.16167},
  year={2025}
}

@article{tian2025interndata,
  title={Interndata-a1: Pioneering high-fidelity synthetic data for pre-training generalist policy},
  author={Tian, Yang and Yang, Yuyin and Xie, Yiman and Cai, Zetao and Shi, Xu and Gao, Ning and Liu, Hangxu and Jiang, Xuekun and Qiu, Zherui and Yuan, Feng and others},
  journal={arXiv preprint arXiv:2511.16651},
  year={2025}
}

@article{zhao2026fdvla,
  title={FD-VLA: Force-Distilled Vision-Language-Action Model for Contact-Rich Manipulation},
  author={Zhao, Ruiteng and Wang, Wenshuo and Ma, Yicheng and Li, Xiaocong and Tay, Francis EH and Ang Jr, Marcelo H and Zhu, Haiyue},
  journal={arXiv preprint arXiv:2602.02142},
  year={2026}
}

@article{OSMO,
      title={OSMO: Open-Source Tactile Glove for Human-to-Robot Skill Transfer}, 
      author={Jessica Yin and Haozhi Qi and Youngsun Wi and Sayantan Kundu and Mike Lambeta and William Yang and Changhao Wang and Tingfan Wu and Jitendra Malik and Tess Hellebrekers},
      journal={arXiv preprint arXiv:2512.08920},
      year={2025}
}

@inproceedings{fan2023arctic,
  title={ARCTIC: A dataset for dexterous bimanual hand-object manipulation},
  author={Fan, Zicong and Taheri, Omid and Tzionas, Dimitrios and Kocabas, Muhammed and Kaufmann, Manuel and Black, Michael J and Hilliges, Otmar},
  booktitle={Proceedings of the IEEE/CVF conference on computer vision and pattern recognition},
  pages={12943--12954},
  year={2023}
}

@inproceedings{chao2021dexycb,
  title={Dex{YCB}: A benchmark for capturing hand grasping of objects},
  author={Chao, Yu-Wei and Yang, Wei and Xiang, Yu and Molchanov, Pavlo and Handa, Ankur and Tremblay, Jonathan and Narang, Yashraj S and Van Wyk, Karl and Iqbal, Umar and Birchfield, Stan and others},
  booktitle={Proceedings of the IEEE/CVF conference on computer vision and pattern recognition},
  pages={9044--9053},
  year={2021}
}

@inproceedings{kwon2021h2o,
  title={{H2O}: Two hands manipulating objects for first person interaction recognition},
  author={Kwon, Taein and Tekin, Bugra and St{\"u}hmer, Jan and Bogo, Federica and Pollefeys, Marc},
  booktitle={Proceedings of the IEEE/CVF international conference on computer vision},
  pages={10138--10148},
  year={2021}
}

@inproceedings{hampali2022keypoint,
  title={Keypoint transformer: Solving joint identification in challenging hands and object interactions for accurate {3D} pose estimation},
  author={Hampali, Shreyas and Sarkar, Sayan Deb and Rad, Mahdi and Lepetit, Vincent},
  booktitle={Proceedings of the IEEE/CVF conference on computer vision and pattern recognition},
  pages={11090--11100},
  year={2022}
}

@inproceedings{hampali2020honnotate,
  title={{HOnnotate}: A method for {3D} annotation of hand and object poses},
  author={Hampali, Shreyas and Rad, Mahdi and Oberweger, Markus and Lepetit, Vincent},
  booktitle={Proceedings of the IEEE/CVF conference on computer vision and pattern recognition},
  pages={3196--3206},
  year={2020}
}

@inproceedings{wang2025hocap,
  title={{HO-Cap}: A capture system and dataset for {3D} reconstruction and pose tracking of hand-object interaction},
  author={Wang, Jikai and Zhang, Qifan and Chao, Yu-Wei and Wen, Bowen and Guo, Xiaohu and Xiang, Yu},
  booktitle={The Thirty-ninth Annual Conference on Neural Information Processing Systems Datasets and Benchmarks Track},
  year={2025}
}

@inproceedings{liu2022hoi4d,
  title={{HOI4D}: A {4D} egocentric dataset for category-level human-object interaction},
  author={Liu, Yunze and Liu, Yun and Jiang, Che and Lyu, Kangbo and Wan, Weikang and Shen, Hao and Liang, Boqiang and Fu, Zhoujie and Wang, He and Yi, Li},
  booktitle={Proceedings of the IEEE/CVF conference on computer vision and pattern recognition},
  pages={21013--21022},
  year={2022}
}

@inproceedings{yang2022oakink,
  title={{OakInk}: A large-scale knowledge repository for understanding hand-object interaction},
  author={Yang, Lixin and Li, Kailin and Zhan, Xinyu and Wu, Fei and Xu, Anran and Liu, Liu and Xie, Sheng and Xu, Kai and Tao, Dacheng},
  booktitle={Proceedings of the IEEE/CVF conference on computer vision and pattern recognition},
  pages={20953--20962},
  year={2022}
}

@inproceedings{zhan2024oakink2,
  title={{OakInk2}: A dataset of bimanual hands-object manipulation in complex task completion},
  author={Zhan, Xinyu and Yang, Lixin and Zhao, Yifei and Mao, Kangrui and Xu, Hanwen and Lin, Zenan and Li, Kailin and Xu, Kai},
  booktitle={Proceedings of the IEEE/CVF conference on computer vision and pattern recognition},
  pages={504--514},
  year={2024}
}

@inproceedings{brahmbhatt2024decaf,
  title={Decaf: Monocular deformation capture for face and hand interactions},
  author={Brahmbhatt, Samarth and Li, Cheng-You and Kim, Heeseung and Zheng, Zerong and Singh, Gurprit and Bernstein, Giljoo and Kim, Taehyun and Kim, Hyeongwoo and Raskar, Ramesh and Sheikh, Yaser},
  booktitle={SIGGRAPH Asia 2024 Conference Papers},
  pages={1--12},
  year={2024}
}

@inproceedings{hassan2019resolving,
  title={Resolving {3D} human pose ambiguities with {3D} scene constraints},
  author={Hassan, Mohamed and Choutas, Vasileios and Tzionas, Dimitrios and Black, Michael J},
  booktitle={Proceedings of the IEEE/CVF international conference on computer vision},
  pages={2282--2292},
  year={2019}
}

@inproceedings{huang2022rich,
  title={Capturing and inferring dense full-body human-scene contact},
  author={Huang, Chun-Hao P and Yi, Hongwei and H{\"o}schle, Markus and Safroshkin, Matvey and Alexiadis, Tsvetelina and Polikovsky, Senya and Scharstein, Daniel and Black, Michael J},
  booktitle={Proceedings of the IEEE/CVF conference on computer vision and pattern recognition},
  pages={13274--13285},
  year={2022}
}

@inproceedings{banerjee2025hot3d,
  title={Hot3d: Hand and object tracking in 3d from egocentric multi-view videos},
  author={Banerjee, Prithviraj and Shkodrani, Sindi and Moulon, Pierre and Hampali, Shreyas and Han, Shangchen and Zhang, Fan and Zhang, Linguang and Fountain, Jade and Miller, Edward and Basol, Selen and others},
  booktitle={Proceedings of the IEEE/CVF Conference on Computer Vision and Pattern Recognition},
  pages={7061--7071},
  year={2025}
}

@inproceedings{moon2020interhand2,
  title={Interhand2. 6m: A dataset and baseline for 3d interacting hand pose estimation from a single rgb image},
  author={Moon, Gyeongsik and Yu, Shoou-I and Wen, He and Shiratori, Takaaki and Lee, Kyoung Mu},
  booktitle={European Conference on Computer Vision},
  pages={548--564},
  year={2020},
  organization={Springer}
}

@article{song2025opentouch,
  title={OPENTOUCH: Bringing Full-Hand Touch to Real-World Interaction},
  author={Song, Yuxin Ray and Li, Jinzhou and Fu, Rao and Murphy, Devin and Zhou, Kaichen and Shiv, Rishi and Li, Yaqi and Xiong, Haoyu and Owens, Crystal Elaine and Du, Yilun and others},
  journal={arXiv preprint arXiv:2512.16842},
  year={2025}
}

@inproceedings{zhao2025egopressure,
  title={Egopressure: A dataset for hand pressure and pose estimation in egocentric vision},
  author={Zhao, Yiming and Kwon, Taein and Streli, Paul and Pollefeys, Marc and Holz, Christian},
  booktitle={Proceedings of the Computer Vision and Pattern Recognition Conference},
  pages={27727--27738},
  year={2025}
}

\newpage
\beginappendix

\section{Hyperparameters}

In our method, we have some hyperparameters that can be tuned during training, as listed in Table~\ref{tab:hyperparameters}. For simulation benchmarks and real robot experiments, we keep the hyperparameters during inference the same as those during training.

\begin{table}[htbp]
    \caption{Hyperparameters for pre-training and post-training.}
    \label{tab:hyperparameters}
    \centering
    \begin{tabular}{lccc}
        \toprule
        \textbf{Hyperparameter} & \textbf{Pre-Training} & \textbf{Post-Training (sim)} & \textbf{Post-Training (real robot)} \\
        \midrule
        \rowcolor{BlockA!20} \multicolumn{4}{l}{\textbf{Training Configuration}} \\
        learning rate & 1e-4 & 1e-4 & 1e-4 \\
        weight decay & 1e-5 & 1e-5 & 1e-5 \\
        warmup ratio & 0.05 & 0.05 & 0.05 \\
        \midrule
        \rowcolor{BlockB!20} \multicolumn{4}{l}{\textbf{Loss Weight}} \\
        action loss weight & 1.0 & 1.0 & 1.0 \\
        tactile loss weight & 1.0 & 1.0 & 1.0 \\

        \midrule
        \rowcolor{BlockA!20} \multicolumn{4}{l}{\textbf{Sequence Configuration and Batch Size}} \\
        max num tokens & 8192 & 8192 & 8192 \\
        expected num tokens & 7680 & 7680 & 7680 \\
        equivalent batch size & 128 & 128 & 128 \\

        \midrule
        \rowcolor{BlockB!20} \multicolumn{4}{l}{\textbf{Image Configuration}} \\
        image size & 448$\times$448 & 224$\times$224 & 224$\times$224 \\
        downsample ratio & 0.5 & 0.5 & 0.5 \\

        \midrule
        \rowcolor{BlockA!20} \multicolumn{4}{l}{\textbf{State \& Action Configuration}} \\
        action chunk size & 32 & 8 & 24 \\
        tactile history size & 4 & 2 & 4 \\
        tactile history stride & 8 & 1 & 4 \\

        \bottomrule
    \end{tabular}
    
\end{table}

\clearpage

\end{document}